\documentclass[conference]{IEEEtran}
\usepackage{times}
\usepackage{amsmath,amssymb}
\usepackage{algorithm}
\usepackage{algpseudocode}
\usepackage{graphicx}

\usepackage[numbers]{natbib}
\usepackage{multicol}
\usepackage[bookmarks=true]{hyperref}
\usepackage{tabularray}
\usepackage{xcolor}
\definecolor{skyblue}{RGB}{0,162,232}

\UseTblrLibrary{booktabs}
\newcounter{tblcell}

\newcommand{\NA}{\textemdash}
\usepackage{caption}
\begin{document}

\title{SID: Sliding into Distribution \\for Robust Few-Demonstration Manipulation}




%
\author{\authorblockN{{Yicheng Ma\textsuperscript{1,2}}\authorrefmark{1},
{Wei Yu\textsuperscript{1,2}}\authorrefmark{1},
Zhian Su\textsuperscript{1,2}, 
Xidan Zhang\textsuperscript{1,2}, 
Huixu Dong\textsuperscript{1,2}\authorrefmark{2}}
\authorblockA{\textsuperscript{1}Grasp Lab, Zhejiang University
\quad\textsuperscript{2}Torch Kernel Co., Ltd.}

\authorblockA{\authorrefmark{1}Equal contribution \quad\authorrefmark{2}Corresponding author, e-mail: huixudong@zju.edu.cn}
}

\maketitle

\begin{abstract}
Generalizing robotic manipulation across object poses, viewpoints, and dynamic disturbances is difficult, especially with only a few demonstrations.
End-to-end visuomotor policies are expressive but data-hungry, while planning and optimization satisfy explicit constraints but do not directly capture the interaction strategies demonstrated by humans.
We propose \emph{Sliding into Distribution} (SID), a structured framework that learns an object-centric motion field from canonicalized demonstrations to iteratively \emph{slide} the system toward the demonstrated manifold and into the reliable operating region of a lightweight egocentric execution policy, mitigating out-of-distribution (OOD) execution.
The motion field provides large corrective motions when far from the demonstration manifold and naturally vanishes near convergence, enabling robust reaching under substantial pose and viewpoint shifts. 
Within the reached regime, an egocentric policy trained with conditioned flow matching performs task-specific manipulation, supported by kinematically consistent point-cloud reprojection augmentation that preserves action--observation consistency. 
Across six real-world tasks, SID achieves approximately 90\% success under OOD initializations with only two demonstrations, with under a 10\% drop under distractors and external disturbances. Overall, SID provides a new paradigm for few-shot manipulation: explicitly managing distribution shift via online distribution recovery. Project website: \url{https://sliding-into-distribution.github.io/}.
\end{abstract}

\IEEEpeerreviewmaketitle

\section{Introduction}
Robust robotic manipulation requires policies that remain reliable under changes in object pose, 
camera viewpoint, and dynamic disturbances. Achieving such robustness from limited demonstrations 
is highly desirable in real deployments but remains challenging~\cite{du2023behaviorretrievalfewshotimitation,xia2024cagecausalattentionenables}. 
In low-coverage datasets, a common failure mode is distribution shift—policies encounter states or observations 
not supported by the demonstrations—often dominating over architectural expressivity limitations~\cite{mandlekar2021matters,mandlekar2020learning}. 
Moreover, a manipulation episode is rarely homogeneous: it typically contains two qualitatively different regimes 
(e.g., the approach phase versus execution).

\begin{figure}[t]
    \centering
    \includegraphics[width=\columnwidth]{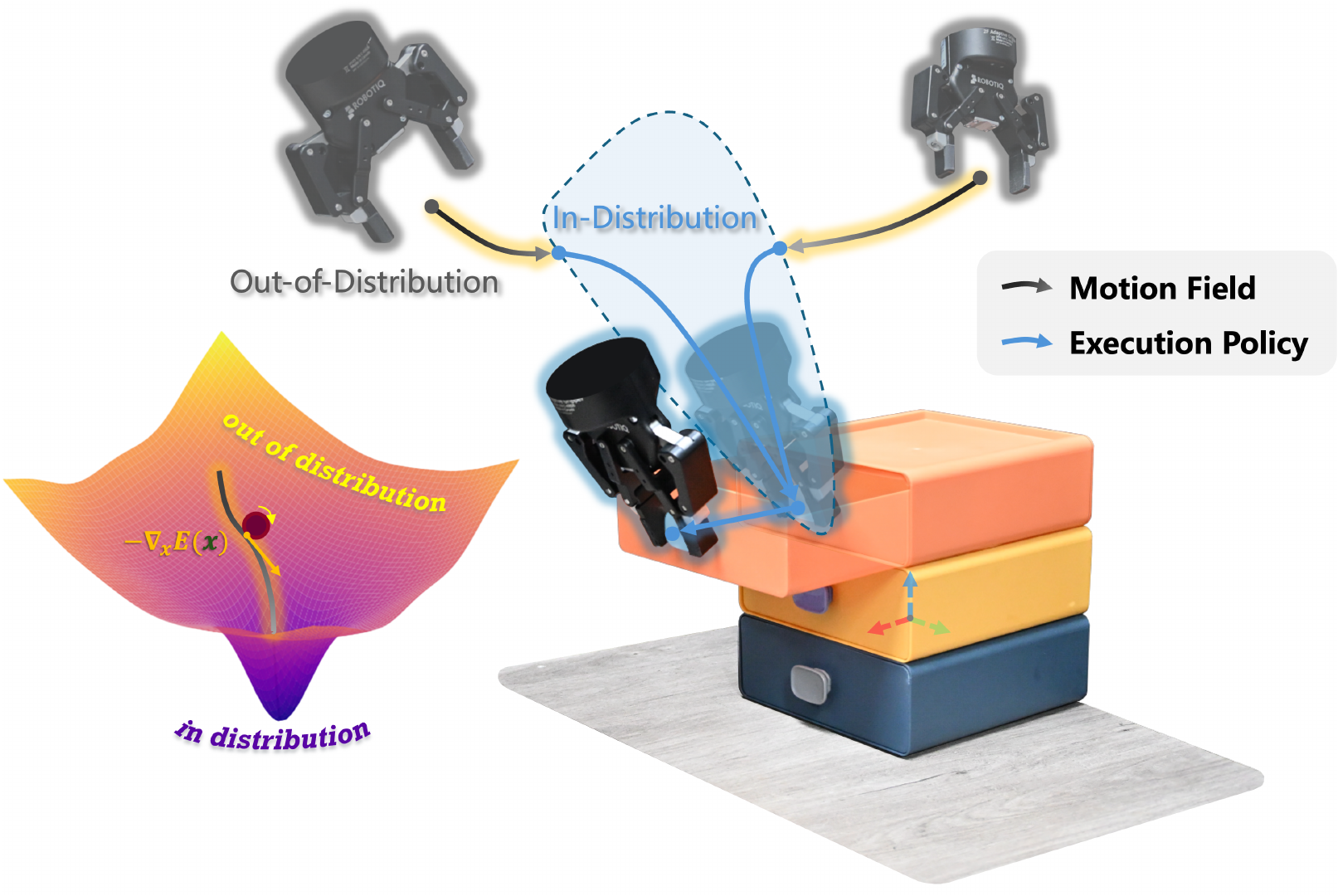}
    \caption{\textbf{SID} combines an object-centric motion field with an egocentric execution policy. 
    From a few demonstrations, the field defines a smooth descent that \emph{slides} OOD states toward the demonstrated support, bringing execution back into the policy's reliable operating region.}
    \label{fig:concept}
\end{figure}

During the approach phase, the robot must recover global geometry to approach a task-relevant target (e.g., a handle or an edge). 
This stage is often underdetermined, admitting multiple feasible approach trajectories and alignments that are consistent with the demonstrations.
During execution, behavior becomes local and interaction-rich, operating in a more constrained regime where it is sensitive to small deviations.
When a single end-to-end visuomotor policy~\cite{levine2016endtoendtrainingdeepvisuomotor,chi2024diffusionpolicyvisuomotorpolicy} 
is trained to cover both regimes with limited demonstrations, it must simultaneously resolve approach-phase ambiguity and execute precise interaction, 
making generalization under pose shifts and perturbations brittle—a failure mode frequently observed in offline imitation with low-coverage data~\cite{mandlekar2021matters}.

Existing learning-based approaches have made steady progress on generalization and data efficiency,
including long-horizon imitation from demonstrations~\cite{mandlekar2020learning,wang2023mimicplay}
and multi-task policy architectures such as transformer-based visuomotor policies~\cite{shridhar2023perceiver}.
These methods often treat the episode as a homogeneous control problem.
Meanwhile, diffusion-based visuomotor policies~\cite{chi2024diffusionpolicyvisuomotorpolicy,ze20243ddiffusionpolicygeneralizable}
can represent complex, multimodal behaviors, but typically require substantial coverage of approach and interaction variations to remain reliable.
Equivariant policy designs~\cite{yang2024equibotsim3equivariantdiffusionpolicy,yang2024equivactsim3equivariantvisuomotorpolicies,wang2024equivariantdiffusionpolicy,zhao2025generalizablehierarchicalskilllearning}
often provide strong OOD robustness when the dominant test-time shift aligns with an underlying symmetry (e.g., rigid $\mathrm{SE}(3)$ transforms).
That said, enforcing equivariance usually comes with architectural commitments that restrict the hypothesis class and introduce additional design/implementation complexity,
potentially trading off some expressive power and ease of deployment for symmetry-consistent generalization.
Coarse-to-fine decomposition is particularly effective in one-shot settings~\cite{johns2021coarse,Dreczkowski_2025},
suggesting that explicitly structuring the control problem can mitigate OOD failures when data are scarce. However, such coarse-to-fine pipelines are often open-loop at execution time and can be sensitive to the initial state being close to the first demonstrated waypoint. 

Building on this observation, we propose SID, a structured framework that combines (i) an object-centric motion field learned from a small set of canonicalized demonstrations and
(ii) an egocentric execution policy. 
We first canonicalize demonstrations, yielding a consistent motion manifold that suppresses scene- and camera-specific variation.
In practice, this canonicalization can be supported by modern 6D pose estimation and tracking~\cite{wen2024foundationposeunified6dpose}
and object-level segmentation~\cite{carion2025sam3segmentconcepts}.

From the resulting manifold, we learn a continuous motion field over monotonic approach-phase segments,
implemented as a gradient-descent-style dynamical system that attracts states toward the demonstrated support.
At test time, integrating the field produces large corrective motions when far from demonstrations and naturally vanishes near convergence,
thereby sliding the system back into the reliable operating region of the execution policy---a process we refer to as \emph{sliding into distribution}.

To make the egocentric policy robust under sparse data without corrupting supervision, we introduce kinematically consistent augmentation through point cloud reprojection: we perturb the end-effector pose, reproject segmented wrist-camera point clouds using fixed hand--eye calibration, and update relative actions consistently
to preserve action--observation validity.
This design is aligned with counterfactual and controllable augmentation paradigms for robot learning~\cite{chen2023genaugretargetingbehaviorsunseen,chen2024semanticallycontrollableaugmentationsgeneralizable,ameperosa2025rocodacounterfactualdataaugmentation},
while explicitly enforcing geometric consistency.
Augmentation is applied only to approach-phase segments, leaving interaction-heavy execution segments unchanged. We also sample a disjoint outer perturbation range to generate OOD observations, which provide negative examples for learning when observations leave the policy's local support.


Finally, within the reached regime, an egocentric execution policy trained via conditioned flow matching performs task-specific manipulation,
building on recent progress in flow-based visuomotor policies~\cite{zhang2024flowpolicyenablingfastrobust,hu2024adaflowimitationlearningvarianceadaptive,sochopoulos2025fastflowbasedvisuomotorpolicies}.
Beyond a simple open-loop handoff from the motion field to the policy, SID also supports a closed-loop variant that uses an auxiliary ID-confidence head, trained from augmentation-induced ID/OOD labels, to trigger field-based re-alignment when observations drift from the policy's local support.
Experiments show that SID achieves strong spatial generalization and robustness to object displacement and disturbances with as few as one to two demonstrations per task.


Overall, we make the following contributions:
\begin{itemize}
    \item We propose SID, a structured framework that combines an object-centric motion field for distribution alignment with an egocentric execution policy for task-specific control, and instantiate it with two inference pipelines: an open-loop field-to-policy handoff and a closed-loop re-alignment variant.
    \item We formulate distribution alignment from few demonstrations as an energy-based dynamical system in object-centric SE(3) space, and learn a smooth motion field that performs gradient-descent-style sliding toward the demonstrated approach manifold.
    \item We introduce a reprojection-based, stage-aware egocentric augmentation scheme that preserves action--observation consistency under pose perturbations and provides ID/OOD labels for the auxiliary ID-confidence head used in the closed-loop variant.
    \item We demonstrate robust generalization across poses, disturbances, cluttered scenes, and long-horizon skill compositions on real-world manipulation tasks with only one to two demonstrations per task.
\end{itemize}

\begin{figure*}[t]
    \centering
    \includegraphics[width=\textwidth]{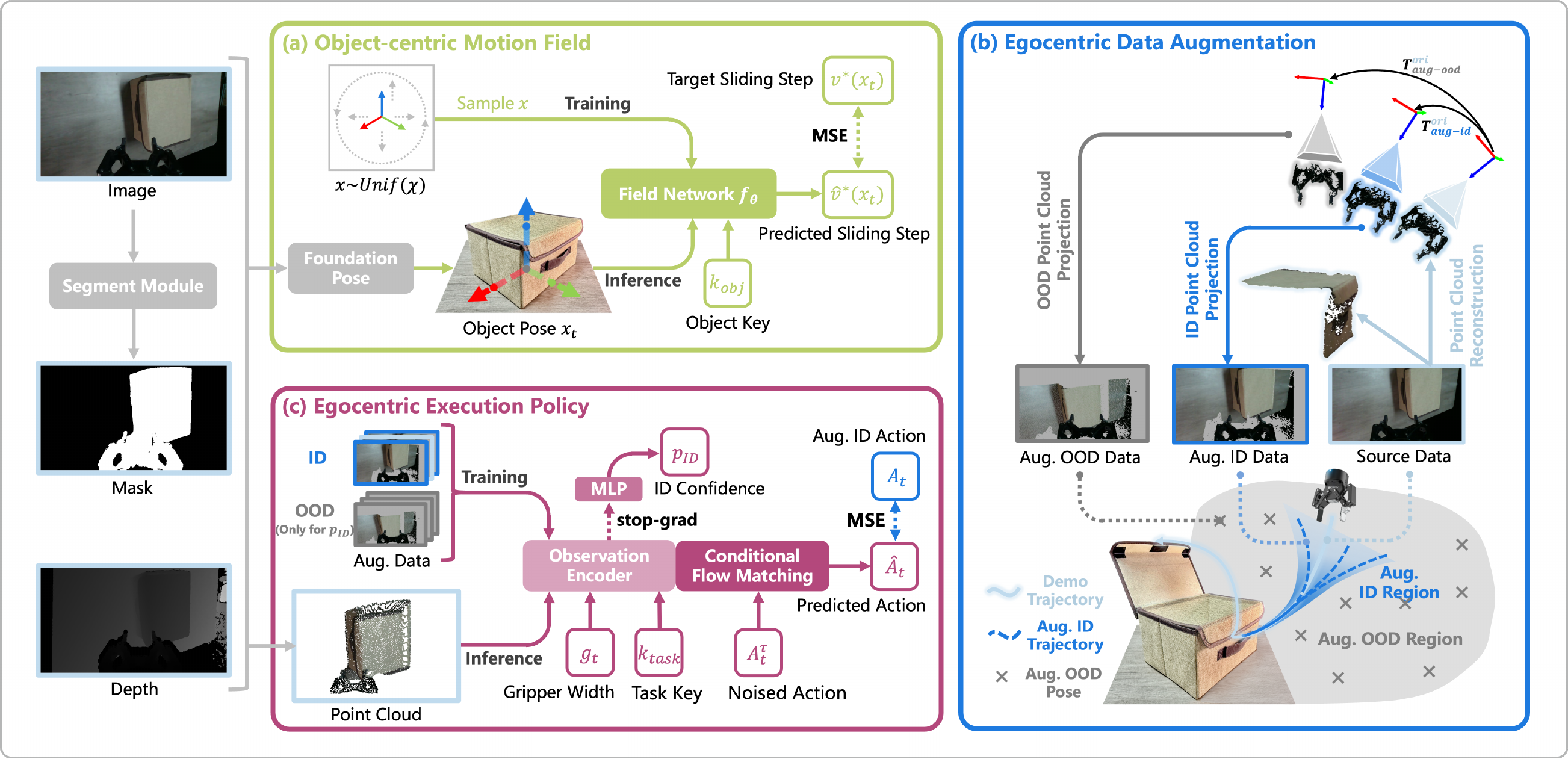}
    \caption{\textbf{SID Overview:} SID comprises three components: (a) an object-centric motion field that predicts sliding steps in object-centric space; (b) an egocentric data augmentation module that generates augmented ID/OOD point-cloud observations via projection and reconstruction; and (c) an egocentric execution policy that predicts actions from point clouds, gripper width, and a task key, with an auxiliary ID-confidence head. }
    \label{fig:overview}
\end{figure*}

\section{Related Work}




\subsection{Visuomotor Policy Learning}
Visuomotor policy learning trains action-generation policies conditioned on visual observations (e.g., RGB, depth, or point clouds), enabling manipulation with visual feedback~\cite{mandlekar2020learning,johns2021coarse,mandlekar2021matters,jang2022bc,wang2023mimicplay,james2022coarse,shridhar2023perceiver}.
Prior work includes end-to-end image-to-action learning~\cite{levine2016endtoendtrainingdeepvisuomotor,levine2016learninghandeyecoordinationrobotic}, temporally-extended control via sequence modeling~\cite{shafiullah2022behaviortransformerscloningk} or action chunking~\cite{zhao2023learningfinegrainedbimanualmanipulation}, and diffusion/flow-based policy formulations for multimodal behaviors~\cite{chi2024diffusionpolicyvisuomotorpolicy,ze20243ddiffusionpolicygeneralizable,zhang2024flowpolicyenablingfastrobust,sochopoulos2025fastflowbasedvisuomotorpolicies,hu2024adaflowimitationlearningvarianceadaptive}.
Recent methods also incorporate 3D representations to capture geometric structure~\cite{ze20243ddiffusionpolicygeneralizable,ke20243ddiffuseractorpolicy,wang2024rise3dperceptionmakes}.
Building on these advances, we adopt an egocentric execution policy (trained with conditioned flow matching) and focus on improving its reliability under sparse supervision.

\subsection{Data-Efficient Manipulation Learning}
Data-efficient manipulation learning aims to reduce demonstration requirements.
Structured policy representations with explicit spatial/geometric inductive biases (e.g., keypoints, transport operators, discretized 3D actions) improve generalization from limited data~\cite{zeng2022transporternetworksrearrangingvisual,shridhar2021cliportpathwaysroboticmanipulation,shridhar2023perceiver,wang2025skilsemantickeypointimitation,manuelli2019kpamkeypointaffordancescategorylevel,Gao_2023}.
Equivariant architectures encode symmetry constraints to improve sample efficiency for pose variations~\cite{yang2024equibotsim3equivariantdiffusionpolicy,yang2024equivactsim3equivariantvisuomotorpolicies,wang2024equivariantdiffusionpolicy,zhao2025generalizablehierarchicalskilllearning}.
Other directions include one-/few-shot adaptation via meta-learning~\cite{finn2017oneshotvisualimitationlearning,duan2017oneshotimitationlearning,zhang2024oneshotimitationlearninginvariance} and synthetic data/augmentation to expand perceptual and behavioral coverage~\cite{zhou2025teachoncelearnoneshot,xue2025demogensyntheticdemonstrationgeneration,chen2023genaugretargetingbehaviorsunseen,chen2024semanticallycontrollableaugmentationsgeneralizable,ameperosa2025rocodacounterfactualdataaugmentation}.
Coarse-to-fine decomposition is also effective in one-/few-shot settings by explicitly structuring the control problem~\cite{johns2021coarse,Dreczkowski_2025,11126904}.
In this spirit, our approach combines a distribution-alignment stage with a task-execution stage to improve data efficiency without increasing the number of demonstrations.

\subsection{Object-centric Policy Learning}
Object-centric policy learning improves robustness by representing goals and behaviors relative to objects rather than the global scene.
Prior work obtains object-aligned representations via detectors~\cite{chapin2025object,li2025language,11063337}, part/affordance segmentation~\cite{qi2025two,geng2023partmanip,li2024unidoormanip,11125003}, or keypoint-based geometric cues~\cite{sundaresan2023kite,tang2025functo,goyal2023rvt,fang2025kalm,qian2024task,li2025bridgevla}, and may further use latent slots~\cite{zhu2023learning} or relational abstractions such as scene graphs~\cite{jiang2024roboexp}.
These object-centric representations are commonly used to parameterize skills, define subgoals, or condition policies in long-horizon tasks.
We similarly leverage object-centric structure, but use it to define and learn a motion field that aligns states back toward the neighborhood of demonstrated trajectories.

\section{Method}
\label{sec:method}

We present \emph{Sliding into Distribution} (SID), a structured manipulation framework illustrated in Fig.~\ref{fig:overview}, designed to generalize from extremely few demonstrations across large object-pose variation, changing wrist viewpoints, and dynamic disturbances.
SID consists of four components:
(i) an \emph{object-centric motion field} $f_\theta$ learned from a small set of canonicalized approach demonstrations,
(ii) an \emph{egocentric data augmentation} module that expands perceptual diversity via kinematically consistent point-cloud reprojection,
(iii) an \emph{egocentric execution policy} $\pi_\theta$ that performs task-specific manipulation once the system is within its reliable operating region, and
(iv) two inference pipeline variants: open-loop and closed-loop.

\subsection{Problem Formulation}
\label{sec:problem_formulation}

We study few-demonstration robot manipulation under workspace-wide variation and external disturbances.
A task is specified by a small demonstration set
$\mathcal{D}_{task}=\{\zeta^{(i)}\}_{i=1}^{M}$ with $M\le 2$,
where each demonstration is a time-indexed sequence
$\zeta^{(i)}=\{(o_t^{(i)}, a_t^{(i)})\}_{t=1}^{T_i}$.
At time $t$, the egocentric observation is $o_t=(P_t, g_t)$, where $P_t$ denotes a wrist-centric colored point cloud reconstructed from the wrist-mounted RGB-D camera, and $g_t\in\mathbb{R}$ is the gripper-width signal.
The action $a_t$ is a per-step relative end-effector motion between neighboring frames (represented in $\mathrm{SE}(3)$).
Our goal is to learn a control system that succeeds under large spatial variation and disturbances while using only $\mathcal{D}_{task}$ as supervision.

\subsection{Object-Centric Motion Field}
\label{sec:field}

\textbf{Why a motion field.}
A major failure mode of egocentric policies under sparse data is distribution shift: 
at test time, the robot may start from observations that are far from the demonstrated 
approach states, causing the policy to operate out of distribution.
SID addresses this by learning an object-centric motion field $f_\theta$ that funnels the system toward the demonstrated approach manifold (see Fig.~\ref{fig:concept}).

\textbf{Canonicalization.}
We use a canonicalization operator $\mathcal{C}$ to extract an object-centric representation $x_t$ and its associated anchor pose $\bar{x}_t$ from an observation:
\begin{equation}
(x_t, \,\bar{x}_t) = \mathcal{C}(o_t;\,\xi_t), 
\qquad x_t \in \mathcal{X}, \,\bar{x}_t \in \Omega.
\end{equation}
Here, $\xi_t$ is a target-object specifier, such as a text prompt. 
Intuitively, the anchor pose specifies the pose of the target object, and thus anchors an object-centered frame in the egocentric frame.
The representation $x_t$ is associated with this anchor, either as an explicit pose or as an object-centric observation, such as segmented RGB or point clouds, associated with the anchor frame.

\begin{figure}[t]
    \centering
    \includegraphics[width=\columnwidth]{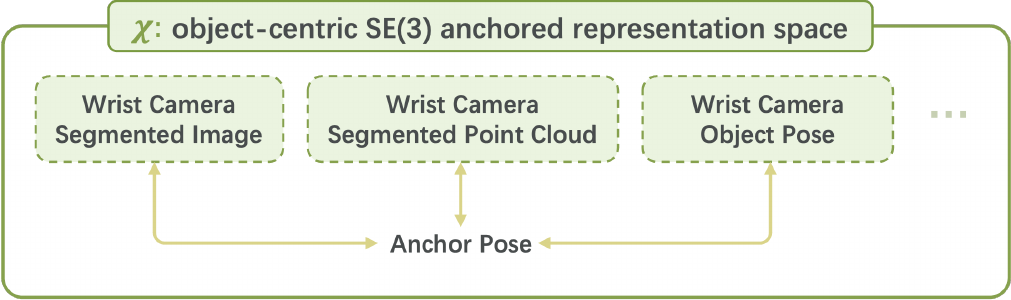}
    \caption{Illustration of the object-centric representation space $\mathcal{X}$ anchored in $\mathrm{SE}(3)$: different instantiations (e.g., wrist-camera segmented images, segmented point clouds, or explicit object poses) share a common \emph{anchor pose} in $\Omega\subset\mathrm{SE}(3)$, enabling pose-aligned distance computations while keeping the network input in the chosen modality.}
    \label{fig:repre}
\end{figure}

In general, $\bar{x}_t \in \Omega \subset \mathrm{SE}(3)$ provides the pose used for pose-aligned geometry, and $\mathcal{X}$ denotes the space of object-centric representations anchored by poses in $\Omega$ 
(see Fig.~\ref{fig:repre}).
In this work, we use a simple and interpretable instantiation by taking $\mathcal{X}=\Omega$, estimating the target-object pose
$T^{c}_{obj}(t)\in \mathrm{SE}(3)$ in the wrist-camera frame, and setting
\begin{equation}
x_t \triangleq \bar{x}_t \triangleq T^{c}_{obj}(t) \in \Omega .
\end{equation}
This pose is used to construct and execute the motion field and to keep the augmentation geometrically consistent, but is not provided as input to the execution policy.

\textbf{Demonstrated approach manifold.}
We restrict the motion field to the monotonic approach portion of each demonstration, where a single-valued corrective behavior toward an approach goal is well defined.
Let $\mathcal{X}_{\text{demo}}=\{x_j\}_{j=1}^{N}\subset\mathcal{X}$ denote the set of object-centric approach states aggregated from $\mathcal{D}_{\text{task}}$, with corresponding anchor poses $\bar{x}_j \in \Omega$.
We use the term \emph{manifold} to refer to the lower-dimensional, task-relevant support formed by the approach-phase states in the demonstrations.

\textbf{Distance-induced sliding field.}
We define a pose-aligned potential $E(x)$ using an $\mathrm{SE}(3)$ distance between anchor poses.
For notational simplicity, we drop the time index and write $x\in\mathcal{X}$ with anchor pose $\bar{x}=[p,q]\in\Omega$, and similarly $\bar{x}_j=[p_j,q_j]\in\Omega$ for each $x_j\in\mathcal{X}_{\text{demo}}$.
We measure similarity with the normalized $\mathrm{SE}(3)$ distance between anchor poses
\begin{equation}
d(x, x_j)\;\triangleq\; d_{\mathrm{SE}(3)}(\bar{x},\bar{x}_j)
= \sqrt{
\left\| \frac{p - p_j}{\sigma_p} \right\|^2 +
\left( \frac{\theta(q, q_j)}{\sigma_r} \right)^2 },
\label{eq:se3_metric}
\end{equation}
where $\theta(q,q_j)$ is the quaternion geodesic angle, and $\sigma_p,\sigma_r$ balance translation and rotation.

We want the field to be well defined not only near demonstrations but everywhere in the bounded pose domain $\Omega$.
To this end, we sample anchor poses $\bar{x}=[p,q]\sim \mathrm{Unif}(\Omega)$ and instantiate the corresponding input $x\in\mathcal{X}$ (the network input) anchored at $\bar{x}$ (e.g., via reconstruction/rendering; when $\mathcal{X}=\Omega$, we simply take $x=\bar{x}$).

We define a potential induced by the demonstrated set,
\begin{equation}
E(x) \triangleq \tfrac{1}{2}\, d(x, \mathcal{X}_{\text{demo}})^2,
\qquad
d(x, \mathcal{X}_{\text{demo}}) \triangleq \min_{x_j\in\mathcal{X}_{\text{demo}}} d(x, x_j),
\end{equation}
and perform a gradient-descent-like update without explicitly computing $-\nabla_x E(x)$.
For any query state $x$, let its nearest demonstrated state be
\begin{equation}
x^*(x)=\arg\min_{x_j\in\mathcal{X}_{\text{demo}}} d(x, x_j).
\end{equation}
We compute the $\mathrm{SE}(3)$ displacement between anchor poses as a 6D twist
\begin{equation}
\tilde{\Delta}(x\!\rightarrow\!x^*)=
\big[p^*-p,\;\omega(q^*,q)\big]^\top \in \mathbb{R}^{6},
\end{equation}
where $\bar{x}^*=[p^*,q^*]\in\Omega$ is the anchor pose of $x^*(x)$, and $\omega(q^*,q)\in\mathbb{R}^3$ is the axis-angle representation of the relative rotation from $q$ to $q^*$.
We then form a unit descent direction
\begin{equation}
\hat{\Delta}(x\!\rightarrow\!x^*)=
\frac{\tilde{\Delta}(x\!\rightarrow\!x^*)}{\|\tilde{\Delta}(x\!\rightarrow\!x^*)\|},
\end{equation}
and define the target sliding step
\begin{equation}
v^*(x)= \eta\!\left(d(x,x^*(x))\right)\,\hat{\Delta}(x\!\rightarrow\!x^*(x)) \in \mathbb{R}^{6}.
\label{eq:target_direction}
\end{equation}
Here $\eta(\cdot)$ is a distance-scheduled step size consistent with the quadratic potential $E(x)=\tfrac12 d^2$, e.g., $\eta(d)\propto d$ with clipping.
Thus, $\hat{\Delta}$ determines the descent direction while $\eta(\cdot)$ determines the magnitude.

\textbf{Learning and execution.}
We parameterize the field as $f_\theta:\mathcal{X}\rightarrow\mathbb{R}^6$ and regress the target steps
\begin{equation}
\mathcal{L}_{\text{field}}
=
\mathbb{E}_{x\sim \mathcal{P}(\mathcal{X})}
\big[
\| f_\theta(x) - v^*(x) \|^2
\big],
\end{equation}
where $\mathcal{P}(\mathcal{X})$ is the induced sampling mixture obtained by sampling anchor poses $\bar{x}\in\Omega$ and instantiating the corresponding inputs $x\in\mathcal{X}$.
At inference time, we obtain the current state $x_t$, query $v_t=f_\theta(x_t)$, and execute the corresponding end-effector command with step size $\alpha$.
We repeat until $\|f_\theta(x_t)\|<\epsilon_{\text{field}}$, indicating that the system has slid near the demonstrated approach manifold, where $\epsilon_{\text{field}}$ is a small threshold indicating that the predicted sliding step magnitude is negligible.

\textbf{Object key for multi-object fields.}
In practice, the motion field additionally conditions on an object key $k_{obj}\in\mathcal{K}$ that identifies the target object (or instance) within the task, enabling a single network to represent fields for multiple objects:
\begin{equation}
f_\theta:\mathcal{X}\times\mathcal{K}\rightarrow\mathbb{R}^6.
\end{equation}

\subsection{Egocentric Data Augmentation}
\label{sec:ego_aug}
\textbf{Egocentric Observation Representation.}
Following the formulation in Sec.~\ref{sec:problem_formulation}, we use point clouds in the camera frame as the observation, yielding an egocentric representation. This egocentric viewpoint is particularly convenient for data augmentation: we can apply rigid transformations---constrained by the hand--eye calibration---to the non-body component of the observation and the end-effector pose, thereby synthesizing new observations and corresponding end-effector states.

\textbf{Data Augmentation via Point Cloud Reprojection.}
To synthesize an egocentric observation under a perturbed viewpoint, we sample a random perturbation from a certain range in the end-effector frame, $T^{ee}_{ee_{\text{aug}}}\in SE(3)$, and convert it to the corresponding camera-frame perturbation using the fixed hand--eye calibration $T^{ee}_{c}$:
\begin{equation}
T^{c}_{c_{\text{aug}}} = (T^{ee}_{c})^{-1}\, T^{ee}_{ee_{\text{aug}}}\, T^{ee}_{c},
\end{equation}
which induces the reprojection transform $T^{c_{\text{aug}}}_{c}=(T^{c}_{c_{\text{aug}}})^{-1}$ mapping points from the original camera frame to the augmented camera frame.

Crucially, to obtain a physically consistent egocentric observation under viewpoint perturbation, we transform only the target-object point cloud while keeping the gripper point cloud unchanged. We apply a segmentation model to $P_t$ to extract the object point cloud $P^{obj}_t$ and the gripper point cloud $P^{grip}_t$, and define their union as $P^{seg}_t = P^{obj}_t \cup P^{grip}_t$.
For each point $\mathbf{p}^c_i(t)\in P^{obj}_t$, we compute
\begin{equation}
\mathbf{p}^{c_{\text{aug}}}_i(t) = T^{c_{\text{aug}}}_{c}\, \mathbf{p}^{c}_i(t),
\end{equation}
where points are represented in homogeneous coordinates.
We leave the gripper points unchanged, i.e.,
\begin{equation}
P^{grip,\,c_{\text{aug}}}_t = P^{grip,\,c}_t,
\end{equation}
and form the augmented segmented observation by
\begin{equation}
P^{seg,\,c_{\text{aug}}}_t \;=\; P^{obj,\,c_{\text{aug}}}_t \,\cup\, P^{grip,\,c_{\text{aug}}}_t.
\end{equation}
while keeping the gripper width unchanged, i.e., $g_t^{\text{aug}}=g_t$.

To preserve action--observation consistency under per-frame perturbations, if we sample $T^{ee}_{ee_{\text{aug}}}(t)$ and $T^{ee'}_{ee'_{\text{aug}}}(t{+}1)$ at consecutive time steps, we update the relative action by
\begin{equation}
a^{\text{aug}}_t
=
\left(T^{ee}_{ee_{\text{aug}}}(t)\right)^{-1}
\, T^{ee}_{ee'}(t\rightarrow t{+}1)\,
T^{ee'}_{ee'_{\text{aug}}}(t{+}1).
\end{equation}
where $T^{ee}_{ee'}(t\rightarrow t{+}1)=a_t$.

Finally, we control the sampling space of $T^{ee}_{ee_{\text{aug}}}$ to generate both in-distribution and out-of-distribution augmentations.
We define ID/OOD relative to the local egocentric support induced by the demonstrations. ID samples are generated from small bounded perturbations around demonstrated end-effector poses and are used, with their geometrically updated actions, to train the execution policy. OOD samples are generated from a disjoint outer perturbation range outside the ID support; they are not used for action-supervised policy training, but serve as negative examples for the auxiliary ID-confidence head.
After augmentation, we attach a binary label $p_{\mathrm{ID}}\in\{0,1\}$ to each sample, where $p_{\mathrm{ID}}{=}1$ indicates ID and $p_{\mathrm{ID}}{=}0$ indicates OOD.

\subsection{Egocentric Execution Policy}

\textbf{Egocentric, Base-Free Policy Interface.}
To promote generalization, the execution policy is defined in a purely egocentric, base-free manner (no global views or base-frame states are used as inputs). This design decouples execution from absolute workspace placement; a detailed generalization analysis is provided in Appendix B.

It conditions only on $o^{seg}_t=(P^{seg}_t,g_t)$ and predicts an action sequence $A_t=[a_t, a_{t+1}, \ldots, a_{t+H-1}] \in \mathbb{R}^{H \times d_a}$,
where $H$ is the action horizon and $d_a$ is the action dimension.
Each per-step action $a_t$ comprises (i) a relative end-effector motion in $SE(3)$, (ii) low-dimensional control scalars for the gripper command, and (iii) a termination signal; the rollout stops when this signal exceeds a threshold.
To support multiple subtasks with a single model, we additionally condition the policy on a task key $k_{task}$ (embedded as a global conditioning vector).

\textbf{Conditional Flow Matching over Action Chunks.}
Following conditional flow matching, we sample a data action chunk $A_t \sim p_{\text{data}}(\cdot \mid o^{seg}_t, k_{task})$
and a base sample $A_t^{0} \sim p_0(\cdot)$ with $p_0=\mathcal{N}(0, I)$.
We define a linear probability path indexed by the flow time $\tau \in [0,1]$:
\begin{equation}
A_t^\tau \;=\; (1-\tau)\,A_t^{0} \;+\; \tau\,A_t,
\qquad \tau \sim \mathcal{U}[0,1].
\end{equation}
The target velocity field along this path is constant:
\begin{equation}
u^*(A_t^\tau \mid A_t, A_t^{0}) \;=\; \frac{dA_t^\tau}{d\tau} \;=\; A_t - A_t^{0}.
\end{equation}
We train a conditional velocity network $u_\theta(A_t^\tau, o_t, k_{task}, \tau)$ by minimizing
\begin{equation}
\mathcal{L}_{\text{CFM}}
=
\mathbb{E}_{A_t,\,A_t^{0},\,\tau}
\left[
\left\|
u_\theta(A_t^\tau, o^{seg}_t, k_{task}, \tau) - (A_t - A_t^{0})
\right\|_2^2
\right].
\end{equation}
At inference time, we sample $A_t^{0} \sim \mathcal{N}(0,I)$ and integrate the learned ODE
$\tfrac{dA_t^\tau}{d\tau} = u_\theta(A_t^\tau, o^{seg}_t, k_{task}, \tau)$ from $\tau=0$ to $\tau=1$ using Euler updates.

\textbf{Network Architecture.}
The velocity network $u_\theta$ encodes the egocentric observation using a PointNet-style visuomotor encoder that jointly embeds the segmented point cloud and gripper width into a conditioning feature.
Conditioned on this feature, the task embedding, and the flow time, an action head predicts the action-space velocity field over the action chunk.

We additionally attach an auxiliary ID-confidence head $c_\phi$ to the shared observation encoder. We write its prediction as $p_{\mathrm{ID}}=c_\phi(o^{seg}_t,k_{task})$, denoting an MLP head applied to the encoder feature produced from $o^{seg}_t$ and $k_{task}$. The output $p_{\mathrm{ID}}\in[0,1]$ estimates whether the current observation lies within the policy's in-distribution region for the current task/stage. ID samples are used for both action-supervised policy training and ID-confidence prediction, while OOD samples serve only as negative examples for this head. We apply stop-gradient from the auxiliary head to the shared encoder to avoid changing the action-prediction representation.

\begin{figure}[t]
    \centering
    \includegraphics[width=\columnwidth]{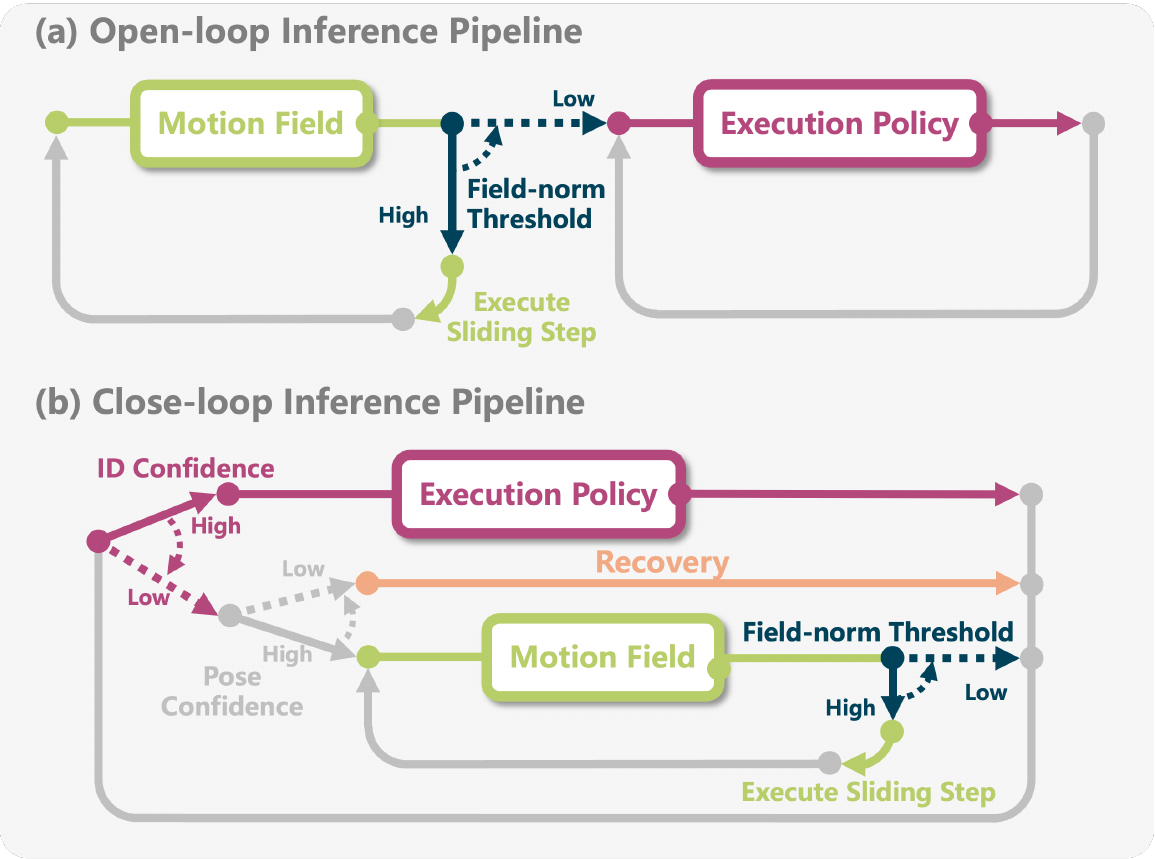}
    \caption{\textbf{Inference pipelines.} (a) Open-loop switches from motion-field sliding (stopped by a field-norm threshold) to policy execution. (b) Closed-loop uses ID/pose confidence to route between policy, motion-field sliding, and recovery.}
    \label{fig:inference}
\end{figure}

\subsection{Inference Procedures of SID}
\label{sec:procedures}

SID supports two inference pipelines: an \emph{Open-Loop Inference Pipeline} and a \emph{Closed-Loop Inference Pipeline}.
The corresponding procedures are summarized in Alg.~\ref{alg:open_loop_sid} and Alg.~\ref{alg:closed_loop_sid}.

\textbf{Open-Loop Inference Pipeline.}
As shown in Alg.~\ref{alg:open_loop_sid}, we first iterate the motion field: the system repeatedly observes $o_t$, canonicalizes to obtain $x_t$, queries $v_t=f_\theta(x_t,k_{obj})$, and executes the sliding step $\alpha\cdot v_t$ until $\lVert f_\theta(x_t,k_{obj})\rVert \le \epsilon_{field}$.
It then switches to the execution policy, executing action chunks $A_t = \pi_\theta(o_t,k_{task})$ until termination.
Termination is determined by the termination signal embedded as one dimension of the $d_a$-dim action, extracted by $\textsc{TermDim}(A_t)$ and thresholded by $\epsilon_{term}$.

\textbf{Closed-Loop Inference Pipeline.}
As shown in Alg.~\ref{alg:closed_loop_sid}, the closed-loop pipeline uses the ID confidence $p_{ID}$ to decide whether to execute the policy.
If $p_{ID}>\epsilon_{ID}$, the policy action $A_t = \pi_\theta(o_t,k_{task})$ is executed and termination is updated via $\textsc{TermDim}(A_t)$.
Otherwise, we canonicalize to obtain $x_t$ and evaluate the pose confidence $p_{pose}=s(x_t)$, where $s(\cdot)$ is the score returned by the foundation pose estimator during pose detection.
If $p_{pose}>\epsilon_{pose}$, we execute sliding steps from the motion field until $\lVert f_\theta(x_t,k_{obj})\rVert \le \epsilon_{field}$; otherwise, we execute a \textsc{Recovery} action and continue the loop. \textsc{Recovery} is a hand-designed fallback that moves the end-effector to a predefined safe observation pose, typically a lifted wrist-camera viewpoint with the target workspace in view. It does not perform task manipulation; instead, it is used to reacquire a more reliable segmentation and pose estimate before re-evaluating $p_{\mathrm{ID}}$ and $p_{\mathrm{pose}}$.

\begin{algorithm}[t]
\caption{Open-Loop Inference Pipeline}
\label{alg:open_loop_sid}
\begin{algorithmic}[1]
\Require Canonicalization operator $\mathcal{C}$, object alignment anchor $\xi_t$, motion field $f_\theta$, execution policy $\pi_\theta$, 
field-norm threshold $\epsilon_{field}$, termination threshold $\epsilon_{term}$, step size $\alpha$, object key $k_{obj}$, task key $k_{task}$.
\Ensure Task termination

\State Observe $o_t$
\State $x_t \leftarrow \mathcal{C}(o_t;\,\xi_t)$
\While{$\lVert f_\theta(x_t, k_{obj}) \rVert > \epsilon_{field}$} \Comment{Motion field}
    \State $v_t \leftarrow f_\theta(x_t, k_{obj})$
    \State \textbf{Execute} sliding step $\alpha \cdot v_t$ 
    \State Observe $o_t$
    \State $x_t \leftarrow \mathcal{C}(o_t;\,\xi_t)$
\EndWhile

\State $\textsc{Terminated} \leftarrow \textsc{False}$
\While{\textbf{not} \textsc{Terminated}} \Comment{Execution Policy}
    \State Observe $o_t$
    \State $A_t \leftarrow \pi_\theta(o_t, k_{task})$
    \State \textbf{Execute} action $A_t$ 
    \State \textsc{Terminated} $\leftarrow$ ($ \textsc{TermDim}(A_t) > \epsilon_{\text{term}}$)
\EndWhile
\end{algorithmic}
\end{algorithm}

\begin{algorithm}[t]
\caption{Closed-Loop Inference Pipeline}
\label{alg:closed_loop_sid}
\begin{algorithmic}[1]
\Require Canonicalization operator $\mathcal{C}$, target-object specifier $\xi_t$, pose score model $s$, pose confidence threshold $\epsilon_{pose}$, motion field $f_\theta$, field-norm threshold $\epsilon_{field}$, execution policy $\pi_\theta$, 
observation encoder with ID confidence head $c_\phi$, ID confidence threshold $\epsilon_{ID}$, termination threshold $\epsilon_{term}$, step size $\alpha$, object key $k_{obj}$, task key $k_{task}$.
\Ensure Task termination
\State $\textsc{Terminated} \leftarrow \textsc{False}$
\While{\textbf{not} \textsc{Terminated}}
    \State Observe $o_t$
    \State $p_{ID} \leftarrow c_\phi$ \Comment{ID confidence}
    \If{$p_{ID} > \epsilon_{ID}$}
        \State $A_t \leftarrow \pi_\theta(o_t, k_{task})$ \Comment{Execution Policy}
        \State \textbf{Execute} action $A_t$
        \State \textsc{Terminated} $\leftarrow$ ($ \textsc{TermDim}(A_t) > \epsilon_{\text{term}}$)
    \Else
        \State $x_t \leftarrow \mathcal{C}(o_t;\,\xi_t)$
        \State $p_{pose} \leftarrow s(x_t)$ \Comment{Pose confidence}
        \If{$p_{pose} > \epsilon_{pose}$}
            \While{$\lVert f_\theta(x_t, k_{obj}) \rVert > \epsilon_{field}$}
                \State $v_t \leftarrow f_\theta(x_t, k_{obj})$
                \State \textbf{Execute} sliding step $\alpha \cdot v_t$ 
                \State Observe $o_t$
                \State $x_t \leftarrow \mathcal{C}(o_t;\,\xi_t)$
            \EndWhile
        \Else
            \State \textbf{Execute} \textsc{Recovery} action
            \State \textbf{continue}
        \EndIf
    \EndIf
\EndWhile
\end{algorithmic}
\end{algorithm}

\begin{figure*}[t]
    \centering
    \includegraphics[width=2\columnwidth]{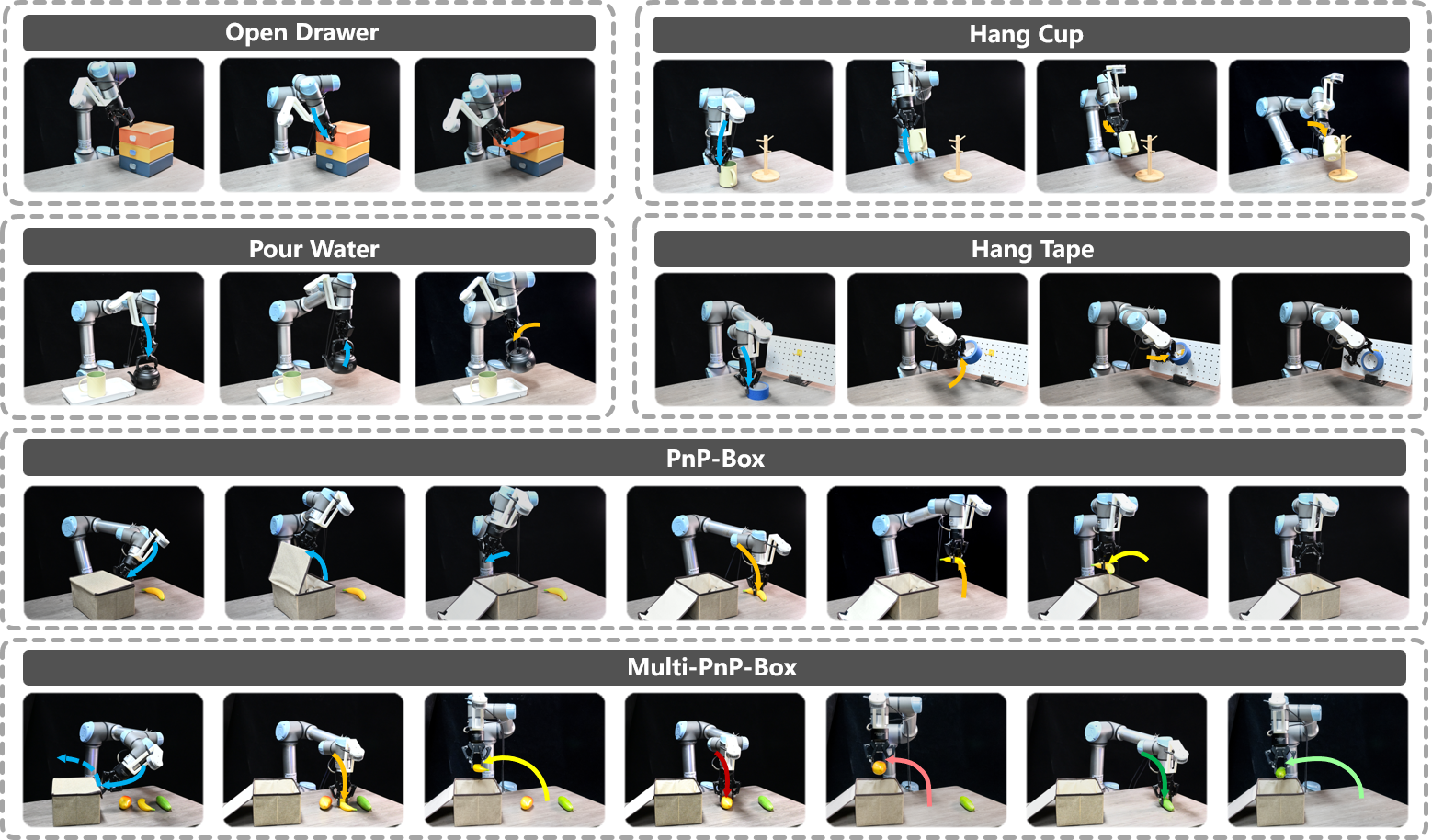}
    \caption{Visualization of the selected tasks. For each task, the robot motion is decomposed into multiple sub-steps, visualized by trajectories whose colors encode the execution order:
    {\color{skyblue}blue} $\rightarrow$ {\color{orange}orange} $\rightarrow$ {\color{yellow}yellow} $\rightarrow$ {\color{red!70!black}dark red} $\rightarrow$ {\color{red!40}light red} $\rightarrow$ {\color{green!60!black}dark green} $\rightarrow$ {\color{green!40}light green}.
    }
    \label{fig:tasks}
\end{figure*}

\section{Experiments}
We design experiments to answer the following questions.
\begin{itemize}
\item \textbf{Q1:} Can SID achieve workspace-wide generalization with at most two demonstrations per task, using only augmentation to expand training samples?
\item \textbf{Q2:} Can SID handle dynamic settings with external disturbances?
\item \textbf{Q3:} Does SID remain reliable in unstructured, cluttered scenes without carefully curated layouts?
\item \textbf{Q4:} Can SID scale to long-horizon manipulation tasks that require sustained, multi-step execution?
\item \textbf{Q5:} Can SID reuse and compose sub-skills across tasks, enabling modular combination of sub-tasks beyond the original demonstrations?
\end{itemize}

\subsection{Experiment Setups}

\subsubsection{Tasks}
To cover the above questions, we evaluate SID on six real-world manipulation tasks: \textsc{Open Drawer}, \textsc{Pour Water}, \textsc{Hang Cup}, \textsc{Hang Tape}, \textsc{PnP-Box}, and \textsc{Multi-PnP-Box}. These tasks cover articulated objects, deformable containers, and cluttered scenes, and require both prehensile and non-prehensile manipulation skills, including approaching, grasping, non-prehensile nudging (to open a fabric container), hanging, pouring, and placing.

\subsubsection{Evaluation Protocol}
Across all experiments, SID uses only two raw demonstrations per task, which are expanded through data augmentation to form 100 training samples. We provide MT3 and Ret-BC with 10 demonstrations per task, as retrieval-based methods benefit from a larger demonstration memory for alignment and interaction. All remaining trainable baselines are trained with 100 demonstrations per task to ensure stable learning and competitive performance under their standard training regime.
We consider five evaluation settings corresponding to Q1--Q5.
First, to test \emph{workspace-wide generalization} (Q1), each task is executed from varied initial object poses and camera viewpoints across the reachable workspace, while limiting data collection to at most two human demonstrations per task.
Second, to assess \emph{robustness to dynamics and disturbances} (Q2),we apply perturbations to four tasks. For a fair comparison, we evaluate all methods except MT3, Ret-BC, and SID with states constrained to the ID region. 
Third, to evaluate \emph{unstructured and cluttered scenes} (Q3), we place distractor objects and vary scene layouts without careful curation.
Fourth, our task suite includes settings that require \emph{long-horizon, multi-step execution} (Q4).
Finally, we include cross-task settings that probe \emph{sub-skill reuse and composition} (Q5) by combining previously seen sub-steps into new task variants. Concretely, we evaluate SID on two novel long-sequence tasks. The first, \textsc{Box Retrieval \& Hanging}, is recomposed from three previously learned skills: \textsc{PnP-Box}, \textsc{Hang Tape}, and \textsc{Hang Cup}. The second, \textsc{Pick New Objects into Box}, is recomposed from \textsc{PnP-Box} and \textsc{Pick Objects} (a multi-object picking skill that involves grasping several objects).

\begin{table*}[t]
\centering
\footnotesize
\renewcommand{\arraystretch}{0.92}
\setcounter{tblcell}{0}
\caption{Results in the static setting under both in-distribution (ID) and out-of-distribution (OOD) evaluations. We report the success rate (\%) over 50 trials for each task. \textit{Training Demos} denotes the number of demonstrations used for training. {SID-O} and {SID-C} denote {SID-open} and {SID-closed}, respectively. $\dagger$ denotes our method.}
\label{tab:id_and_ood}

\begin{tblr}{
  width=\linewidth,
  colspec={X[1,c] X[1.2,c] *{12}{X[1,c]}}, 
  cells={c,m},
  column{1}={mode=text},
  column{2}={mode=text},
  colsep=2pt,
  rowsep=1pt,
  hline{1,3,Z}={-}{0.16em},
  hline{2}={3-Z}{},
  vline{2,3}={1-Z}{},
  vline{5,7,9,11,13}={1-Z}{},
}

\SetCell[r=2]{c} Methods
  & \SetCell[r=2]{c} Training Demos
  & \SetCell[c=2]{c} Open Drawer
  & & \SetCell[c=2]{c} Pour Water
  & & \SetCell[c=2]{c} Hang Tape
  & & \SetCell[c=2]{c} Hang Cup
  & & \SetCell[c=2]{c} PnP-Box
  & & \SetCell[c=2]{c} Multi-PnP-Box
  & \\
& & ID & OOD  & ID & OOD  & ID & OOD  & ID & OOD  & ID & OOD  & ID & OOD \\

ACT        & 100 & 82 & 0 & 68 & 0 & 62 & 0 & 52 & 0 & 56 & 0 & 18 & 0 \\
DP3        & 100 & 86 & 10 & 74 & 6 & 56 & 2 & 46 & 4 & 72 & 12 & 24 & 0 \\
$\pi_{0.5}$      & 100 & 92 & 14 & 82 & 4 & 84 & 6 & 78 & 6 & 76 & 10 & 64 & 0 \\

Ret-BC     & 10   & \NA & 56 & \NA & 54 & \NA & 36 & \NA & 62 & \NA & 44 & \NA & 36 \\
MT3        & 10   & \NA & 76 & \NA & 78 & \NA & 46 & \NA & 74 & \NA & 58 & \NA & 52 \\

SID-O$^{\dagger}$   & 2   & \NA & 90 & \NA & 86 & \NA & 82 & \NA & 86 & \NA & 84 & \NA & 82 \\
SID-C$^{\dagger}$ & 2   & \NA & \textbf{92} & \NA & \textbf{90} & \NA & \textbf{88} & \NA & \textbf{90} & \NA & \textbf{92} & \NA & \textbf{86} \\

\end{tblr}
\end{table*}

\subsubsection{Baselines}
We compare our method against five representative and strong baselines.
(1) \textbf{$\boldsymbol{\pi_{0.5}}$}~\cite{intelligence2025pi05visionlanguageactionmodelopenworld}, a vision-language-action (VLA) model trained via co-training on heterogeneous data sources across robot platforms and language-conditioned behaviors to generalize to new environments.
(2) \textbf{MT3}~\cite{Dreczkowski_2025}, a fully retrieval-based decomposition method that performs both alignment and interaction via demonstration retrieval at test time.
(3) \textbf{Ret-BC}~\cite{Dreczkowski_2025}, a hybrid decomposition baseline that uses retrieval-based alignment followed by behavioral cloning (BC) for interaction.
(4) \textbf{3D Diffusion Policy (DP3)}~\cite{ze20243ddiffusionpolicygeneralizable}, a point-cloud diffusion policy that integrates a simplified 3D encoder with robot proprioception for generalizable manipulation.
(5) \textbf{Action Chunking Transformer (ACT)}~\cite{zhao2023learningfinegrainedbimanualmanipulation}, a transformer-based policy that predicts temporally extended action chunks from visual observations.
All baselines are evaluated under identical experimental conditions for a fair comparison.

\subsubsection{Metrics}
We report \textbf{Success Rate} for all tasks. Additionally, we report \textbf{Average Length}~\cite{mees2022calvinbenchmarklanguageconditionedpolicy} for long-horizon tasks, which measures execution progress as the number of stages successfully completed. 
Formally, for a $K$-stage task, each trial is assigned $l \in \{0,1,\dots,K\}$, where $l$ is the index (count) of the last successfully completed stage; $l=K$ indicates full task completion. 


\subsection{Results Comparison}

\textbf{(Q1)SID achieves workspace-wide generalization under OOD initializations with at most two demonstrations per task, consistently outperforming strong baselines.}
Results are summarized in Table~\ref{tab:id_and_ood}. Under the OOD setting, all baselines other than MT3 and Ret-BC suffer pronounced performance degradation. By comparison, both SID-open and SID-closed remain robust in OOD, and outperform these baselines not only in OOD but also relative to their ID performance.

We further observe that SID consistently surpasses MT3 and Ret-BC across tasks. We attribute this improvement to SID’s ability to update its motion field online using the latest egocentric observations while transitioning from OOD regions toward the ID manifold, enabling more accurate re-entry into the training distribution. In contrast, MT3 and Ret-BC perform trajectory alignment based on object poses estimated from a single first-frame global observation, making them sensitive to pose estimation errors that can propagate to downstream manipulation accuracy.
We also observe from Table~\ref{tab:id_and_ood} that SID-closed consistently outperforms SID-open. We attribute this gain to the fact that SID-closed corrects distribution shifts not only during the approach phase, but also continues to mitigate distribution drift in the post-contact phase, where inaccuracies in policy actions can otherwise accumulate and push the system further away from the training distribution.

Overall, these results provide clear evidence that SID delivers strong workspace-wide generalization with minimal demonstrations: OOD-to-ID motion and augmentation enable effective learning from only two demos per task, and the closed-loop variant further improves robustness by continuously correcting distribution drift throughout execution.

\textbf{(Q2) SID performs robustly under external disturbances by correcting deviations online.}
As shown in Table~\ref{tab:dynamic_task}, $\pi_{0.5}$ exhibits the strongest dynamic robustness across all baselines, which we attribute to its closed-loop hierarchical policy with re-planning from fresh observations and its continuous action-expert design. MT3 and Ret-BC suffer substantial performance drops in dynamic scenarios; both degrade severely when perturbations are present, because retrieval-based methods perform alignment only w.r.t.\ the first observation frame and do not update the alignment online. Ret-BC performs relatively better, likely because the BC-based execution stage provides additional robustness.


SID-open is more robust to \emph{disturbances} since its motion field is continuously updated from the latest observations in approach phases, guiding the robot back toward in-distribution states. However, SID-open can still degrade under stronger disturbances in execution phases: once the disturbance pushes observations outside the in-distribution region, the execution policy alone often fails to recover. In contrast, SID-closed remains robust across both approach and execution phases; its OOD detection module provides a closed-loop safeguard that effectively handles OOD events induced by disturbances.

\begin{table}[t]
\centering
\footnotesize
\renewcommand{\arraystretch}{0.92}

\caption{Results in the dynamic setting. We report the success rate (\%) over 50 trials per task (aggregated over perturbation cases).}
\label{tab:dynamic_task}

\begin{tblr}{
  width=\linewidth,
  colspec={X[1.6,c] *{7}{X[c]}}, 
  cells={c,m},
  column{1}={mode=text},
  colsep=2pt,
  rowsep=1pt,
  hline{1,Z}={-}{0.16em},
  vline{2}={1-Z}{},
  hline{2}={-}{0.11em},
}

Task
& ACT
& DP3
& $\pi_{0.5}$
& Ret-BC
& MT3
& SID-O$^{\dagger}$
& SID-C$^{\dagger}$ \\

Hang Tape   & 6 & 14 & 80 & 12 & 4 & 44 & \textbf{84} \\
Hang Cup    & 2 & 8 & 68 & 4 & 0 & 36 & \textbf{88} \\
PnP-Box     & 2 & 10 & 74 & 8 & 2 & 52 & \textbf{86} \\
Pour Water  & 4 & 10 & 78 & 6 & 2 & 40 & \textbf{82} \\

\end{tblr}
\end{table}

\textbf{(Q3) SID remains reliable in unstructured, cluttered scenes.}
As shown in Table~\ref{tab:clutter_task}, MT3, Ret-BC, and both SID variants maintain strong performance under cluttered, unstructured layouts. We attribute this robustness primarily to the use of vision foundation models: SAM2/SAM3 provide explicit target-object segmentation, which filters out background clutter and distractors and yields more stable, target-centric visual inputs for downstream planning and control. As a result, these methods are less sensitive to layout curation and remain reliable even with non-canonical object arrangements. In contrast, the remaining baselines degrade markedly in clutter, since their visual representations are more easily dominated by irrelevant objects and occlusions when explicit target segmentation is unavailable, leading to mis-localization and cascading execution errors.

\begin{table}[t]
\centering
\footnotesize
\renewcommand{\arraystretch}{0.92}
\setcounter{tblcell}{0}

\caption{Results in cluttered environments with added distractor objects. For each task, we report the success rate (\%) over 50 trials per task for each method.}
\label{tab:clutter_task}

\begin{tblr}{
  width=\linewidth,
  colspec={X[1.5,c] *{8}{X[1,c]}}, 
  cells={c,m},
  column{1}={mode=text},
  colsep=2pt,
  rowsep=1pt,
  hline{1,Z}={-}{0.16em},   
  hline{2}={-}{0.11em},
  vline{2}={1-Z}{},           
}

Task & ACT & DP3 & $\pi_{0.5}$ & Ret-BC & MT3 & SID-O$^{\dagger}$ & SID-C$^{\dagger}$ \\

\SetCell[r=1]{c} Hang Cup
& 42 & 24 & 72 & 58 & 72 & 86 & \textbf{88} \\

\SetCell[r=1]{c} PnP-Box
& 44 & 28 & 74 & 40 & 54 & 84 & \textbf{90} \\

\end{tblr}
\end{table}








\textbf{(Q4) SID can handle long-horizon manipulation tasks that require sustained, multi-step execution.}
As shown in Table~\ref{tab:id_and_ood}, SID achieves consistently strong performance on two long-horizon tasks with more than three sequential stages, outperforming the baselines. We attribute this advantage to SID's decomposition design: the motion field facilitates transitions between the subtask-specific state distributions of consecutive stages, while the execution policy only needs to operate within each subtask-relevant, narrow distribution. This separation substantially reduces the learning burden of the neural policy, enabling reliable multi-stage execution with a relatively compact model. Overall, the results suggest that decomposing long-horizon manipulation into distribution-aware subtask transitions is key to sustained, multi-step control.

\begin{figure}[t]
    \centering \includegraphics[width=\columnwidth]{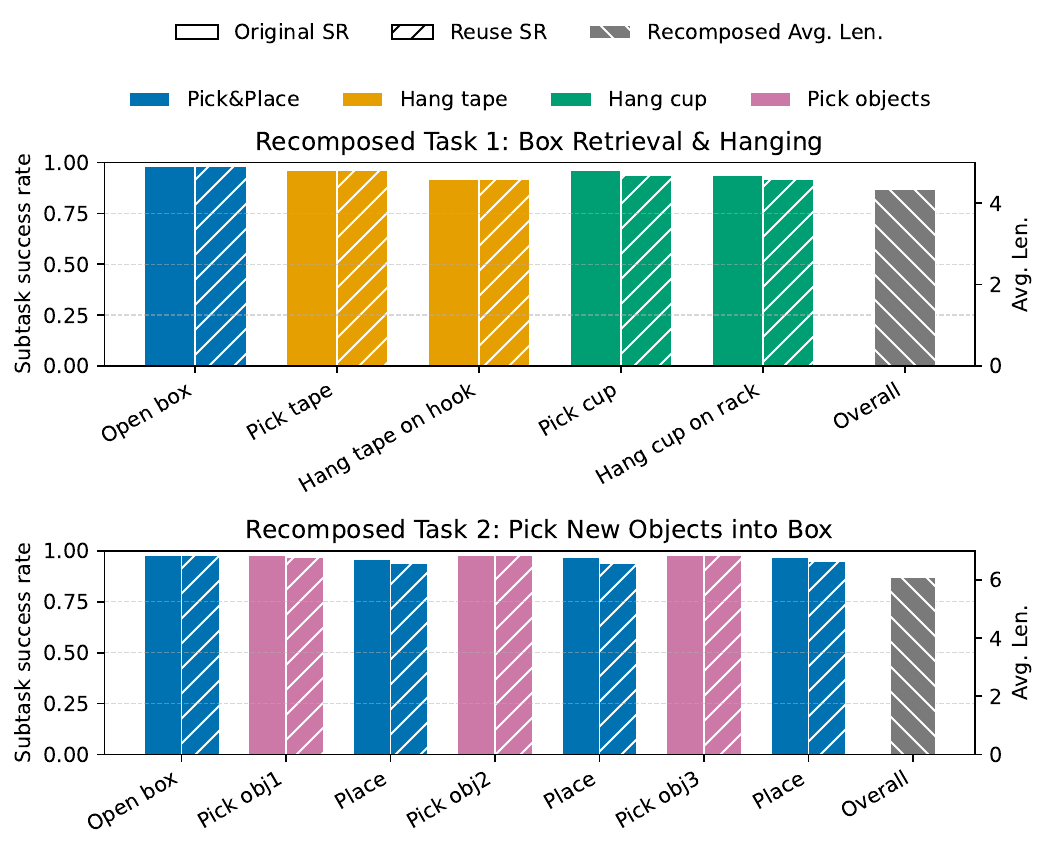}
   \caption{Results on recomposed tasks. We report the success rate for each sub-skill and the Average Length for the overall recomposed tasks, which counts the number of stages successfully completed.}

    \label{fig:recompose}
\end{figure}

\textbf{(Q5) SID can reuse and compose sub-skills across tasks, enabling modular combination of sub-tasks beyond the original demonstrations.}
Since both components in SID are lightweight, we can load multiple previously trained SID models when solving a new task and invoke them stage-by-stage, enabling practical skill reuse and re-composition. 
As shown in Fig.~\ref{fig:recompose}, SID achieves competitive performance on both recomposed tasks by reusing pre-trained sub-skill models and invoking them on demand at each stage. This result highlights SID's ability to recombine existing skills to solve novel long-horizon sequences without retraining an end-to-end policy.

Meanwhile, the per-stage success rates in Fig.~\ref{fig:recompose} indicate a consistent gap between \emph{reuse} and the original single-skill success rates. The degradation is most pronounced for \texttt{Pick tape} and \texttt{Pick cup}, because grasping in the original tasks is performed without occlusions, whereas in the recomposed setting the box introduces additional clutter and partial occlusion that interferes with perception and approach. We also observe reduced performance in the \texttt{Place} stages: since the manipulated objects differ from those seen in the original \textsc{PnP-Box} demonstration, the resulting distribution shift further degrades placement quality.

In summary, SID enables effective skill reuse and modular re-composition for new long-horizon tasks, while the remaining performance gap is largely explained by occlusion-induced grasping difficulty and distribution deviations introduced by novel objects and interactions.

\section{Conclusions and Limitations}
We presented \emph{Sliding into Distribution} (SID), a few-demonstration manipulation framework that mitigates distribution shift by pairing an object-centric motion field for online alignment with an egocentric execution policy trained via conditioned flow matching.
SID learns a smooth descent field from canonicalized approach demonstrations to \emph{slide} OOD states back toward the demonstrated manifold, and improves data efficiency through kinematically consistent point-cloud reprojection augmentation.
Across real-world tasks, SID shows strong workspace-wide generalization and robustness to disturbances with only one to two demonstrations, while the closed-loop variant further uses an auxiliary ID-confidence signal to monitor policy-side distribution drift and route control back to field-based re-alignment or recovery when needed.
SID also supports skill reuse and composition for modular long-horizon manipulation.

\textbf{Limitations:}
\emph{Pose estimation dependency.}
SID relies on an off-the-shelf 6D pose estimator; when pose estimates are unreliable, such as for transparent objects or rapid motion, the motion field can degrade.
Future work could reduce this dependency by conditioning on object-centric inputs such as segmented point clouds or RGB.
\emph{Limited scene awareness and approach feasibility.}
Our method does not explicitly model obstacles or collision constraints, and its motion-field alignment assumes that the demonstrated approach manifold remains observable and reachable.
Performance may degrade when feasible approach regions are occluded, blocked, or constrained by cluttered environments.
\emph{Limited cross-object generalization.}
We do not yet demonstrate strong generalization to unseen objects or categories; more object-agnostic representations are a natural next step.

\section*{Acknowledgments}
This work was supported in part by the National Natural Science Foundation of China Youth Program under Grant No. 52305037, Zhejiang Provincial Natural Science Foundation of China under Grant No. LD26E050001, the "Pioneer" and "Leading Goose" R\&D Programs of Zhejiang Province under Grants  No. 2025C01072.

\bibliographystyle{plainnat}
\bibliography{references}

\clearpage




\pdfinfo{
   /Author (Homer Simpson)
   /Title  (Robots: Our new overlords)
   /CreationDate (D:20101201120000)
   /Subject (Robots)
   /Keywords (Robots;Overlords)
}
\setcounter{figure}{6}
\setcounter{table}{3}

\section*{Appendix A}
\label{sec:appendixA}

\subsection{Hardware Setup}
\begin{figure}[h]
    \centering
    \includegraphics[width=0.95\linewidth]{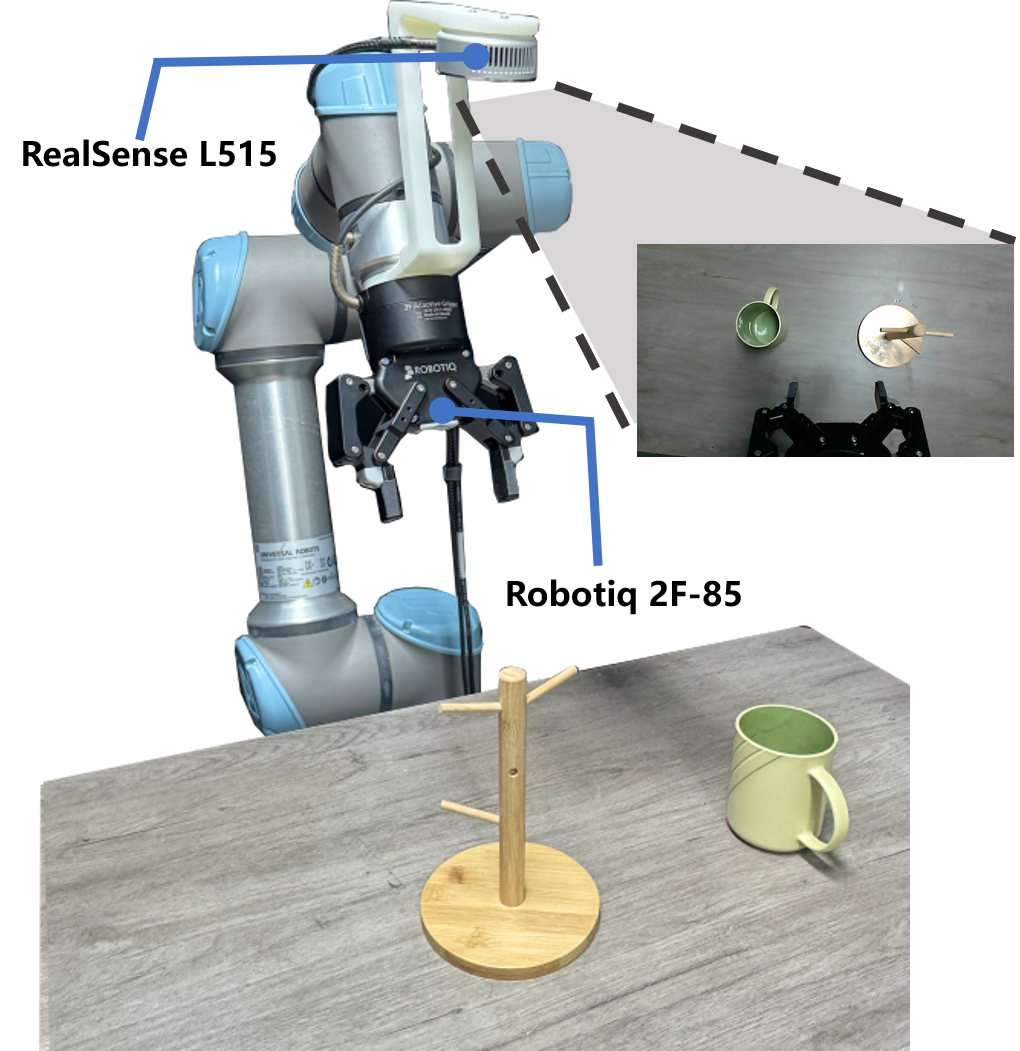}
    \caption{Hardware setup used in our experiments. The platform consists of a UR5e arm with a Robotiq 2F-85 gripper and an Intel RealSense L515 RGB-D camera mounted for egocentric  observations. The inset shows a representative camera view used during execution.}
    \label{fig:hardware}
\end{figure}
As shown in Fig.~\ref{fig:hardware}, our experimental platform consists of a 6-DoF UR5e robotic manipulator equipped with a Robotiq 2F-85 parallel gripper and an Intel RealSense L515 RGB-D camera. 
The RealSense L515 is mounted to provide egocentric visual observations of the workspace, while the Robotiq 2F-85 gripper is attached to the end-effector of the UR5e. The camera is calibrated with respect to the robot end-effector frame, enabling pointcloud projection and unprojection under end-effector frame. Additionally, we select the camera mounting location to ensure that, under a suitable end-effector pose, the camera provides near-complete coverage of the workspace while keeping the gripper within view.

\subsection{Task Description}
The specific task descriptions of six main tasks are as follows: 

\begin{itemize}
    \item \textsc{Open Drawer}: The robot must open the topmost layer of a three-layer drawer, requiring precise positioning and manipulation.
    
    \item \textsc{Pour Water}: The task consists of two steps: first, the robot grasps a teapot, and then it pours water into a cup by tilting the teapot.
    
    \item \textsc{Hang Cup}: In this task, the robot grasps a cup and hangs it on a rack, requiring accurate grasping and placement.
    
    \item \textsc{Hang Tape}: The robot must grasp a roll of tape and hang it on a hook, involving careful handling and placement of the tape.
    
    \item \textsc{PnP-Box}: The robot opens a canvas box, grasps an object from a table, and places it inside the box, testing its ability to handle and place objects.
    
    \item \textsc{Multi-PnP-Box}: The robot first opens a canvas box, then sequentially grasps multiple objects from a table and places them into the box in a predefined order, testing its ability to execute repeated pick-and-place operations and maintain long-horizon, multi-stage coordination.
\end{itemize}

\begin{figure*}[t]
    \centering
    \includegraphics[width=0.95\linewidth]{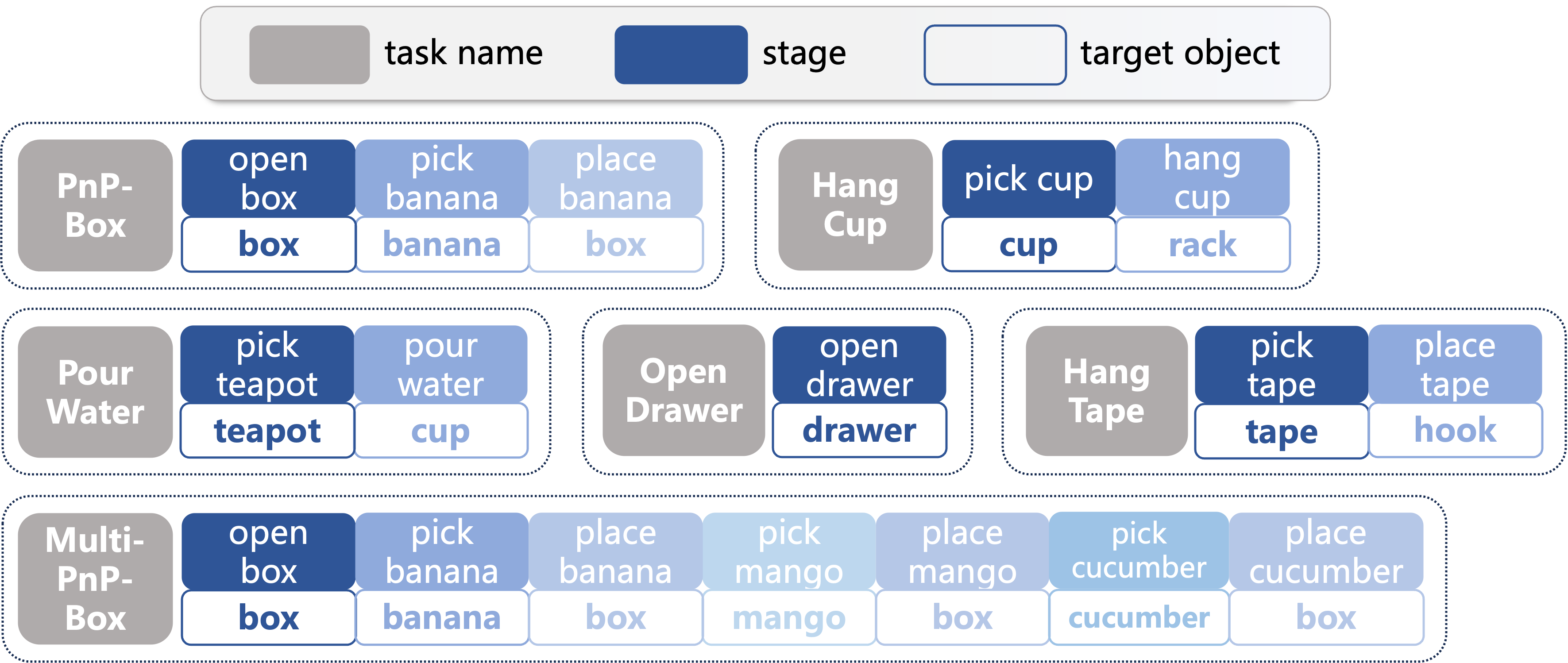}
    \caption{Stage-wise decomposition of the six main tasks, along with the target object associated with each stage.}
    \label{fig:task_decomposition}
\end{figure*}

\subsection{OOD and ID Settings for Baselines}
\label{sec:ood_id_settings}

\begin{figure*}[t]
    \centering
    \includegraphics[width=\textwidth]{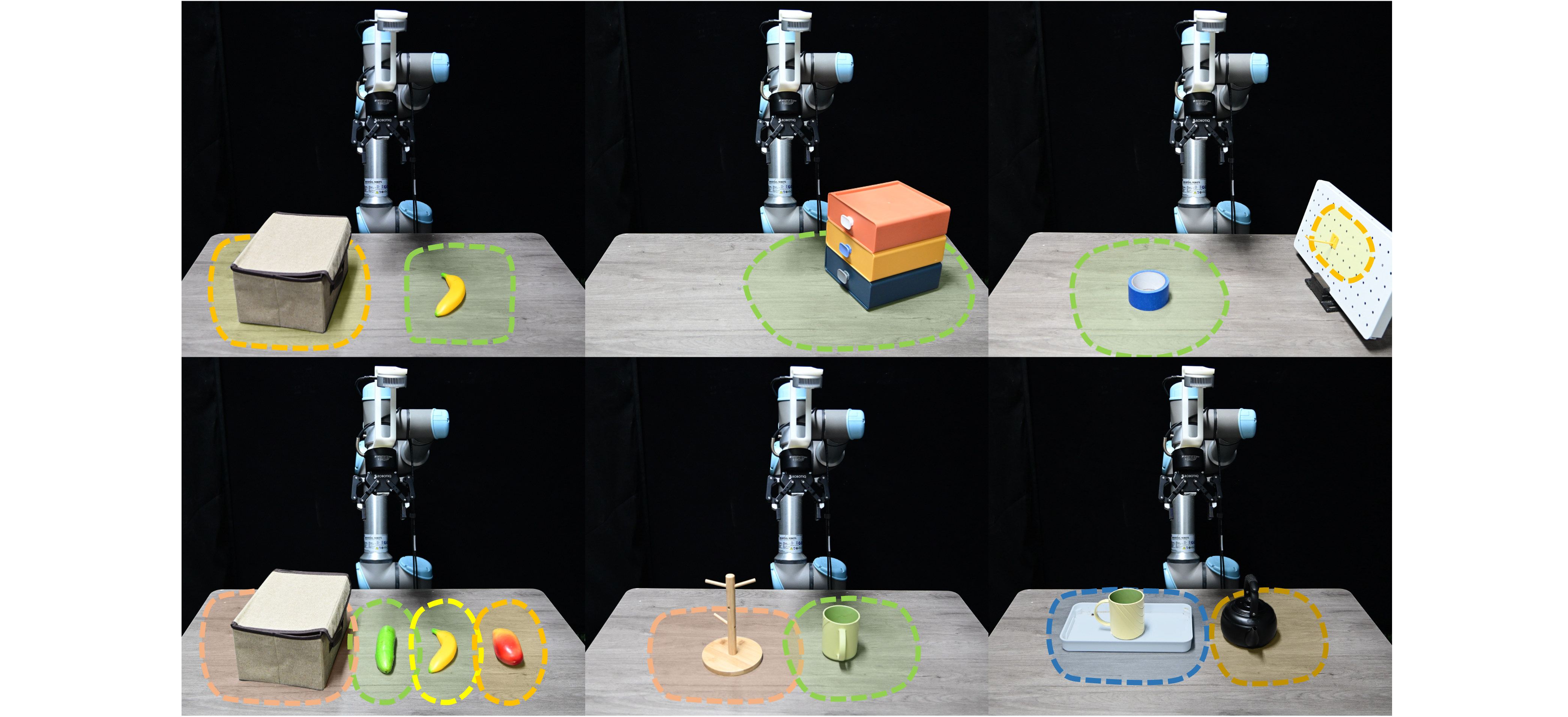}
    \caption{\textbf{ID settings (baselines).} Dashed regions indicate the in-distribution (ID) object-placement areas used to collect training data for imitation baselines. OOD evaluations place the relevant object(s) outside these regions while keeping the rest of the setup unchanged.}
    \label{fig:id_settings}
\end{figure*}

We define in-distribution (ID) and out-of-distribution (OOD) settings \emph{specifically for training and evaluating imitation baselines}.
The ID region of each task is the object-placement area from which we collect the baseline training dataset.
Concretely, the dashed areas in Fig.~\ref{fig:id_settings} visualize the ID regions (one or more regions per task, depending on the scene).

OOD settings are created by placing the relevant object(s) \emph{outside} the corresponding ID region(s), while keeping the rest of the setup unchanged.
Unless otherwise stated, all baseline results reported under ID/OOD follow this protocol.

\subsection{Ablation Experiments}
We conduct two ablation studies to quantify the impact of egocentric design choice and data augmentation in SID. 
The tasks, hardware setup, and evaluation metrics follow the main experiments.
\subsubsection{Ablation 1: Effect of Egocentric Data Augmentation}
\paragraph{Setting.}
We compare two training settings: (i) \textbf{w/ ego-aug}, where we apply egocentric data augmentation to expand the dataset from 2 demonstrations to 50 demonstrations, and (ii) \textbf{w/o ego-aug}, where we train using only the original 2 demonstrations without augmentation.
All other training and inference configurations are kept identical.

\paragraph{Results.}
\begin{table}[t]
\centering
\small
\setlength{\tabcolsep}{6pt}
\begin{tabular}{lccc}
\toprule
\textbf{Task} & \textbf{w/ ego-aug} & \textbf{w/o ego-aug} & $\Delta$ \\
\midrule
Open Drawer & 92 & 18 & 74 \\
Hang Tape & 88 & 4 & 84 \\
Hang Cup & 90 & 0 & 90 \\
Pour Water & 90 & 0 & 90 \\
PnP-Box & 92 & 0 & 92 \\
Multi-PnP-Box & 86 & 0 & 86 \\
\midrule
\textbf{Avg.} & 89.7 & 3.7 & 86 \\
\bottomrule
\end{tabular}
\caption{Ablation on egocentric data augmentation. We report per-task success rates (\%) under identical settings, with and without egocentric data augmentation. $\Delta$ denotes the absolute improvement.}
\label{tab:ablation_ego_aug}
\end{table}
Table~\ref{tab:ablation_ego_aug} reports the per-task success rates with and without egocentric data augmentation. 
Overall, enabling ego-aug substantially improves performance across all tasks, with the largest gains observed on long-horizon tasks.

\paragraph{Analysis.}
We attribute the improvements to two complementary effects. 
First, while the motion field is designed to steer the camera toward in-distribution viewpoints, its pose estimation module inevitably incurs error. 
Without augmentation, the training distribution of egocentric observations is extremely narrow, leaving little tolerance for such errors and making it difficult for the system to reliably enter (and remain in) the in-distribution region. 
Egocentric augmentation artificially broadens the support of the training distribution, effectively enlarging the ``capture range'' of the motion field such that, even with moderate pose errors, the executed viewpoint can still fall within the learned distribution.

Second, the broadened observation distribution also improves the robustness of the execution policy itself. 
By exposing the policy to a wider range of camera poses and the resulting appearance/occlusion variations (e.g. partial visibility of the gripper/object), augmentation reduces overfitting to specific viewpoints and increases tolerance to observation noise and small geometric misalignments. 
These effects become particularly important for long-horizon tasks, where small errors compound over time and the visual input can drift away from the nominal training conditions.

\subsubsection{Ablation 2: Egocentric vs.\ Fixed Camera Viewpoint}
\paragraph{Setting.}
To isolate the effect of observation viewpoint, we collect demonstrations with the egocentric camera and an external fixed camera mounted simultaneously.
This yields paired observations for each timestep from identical trajectories, ensuring that the training data are matched and the only difference between settings is the camera viewpoint.
We use the same set of 50 demonstrations for both settings and train two policies with identical architectures and hyperparameters: one using egocentric RGB-D observations and the other using fixed-camera RGB-D observations.
To avoid confounding factors, we do not apply any data augmentation in this comparison.
Moreover, to factor out the influence of other components in our pipeline, the motion field is trained and queried using egocentric observations in both settings (including the fixed-camera setting); only the execution policy receives fixed-camera observations in that setting.

\paragraph{Results.}
\begin{table}[t]
\centering
\small
\setlength{\tabcolsep}{6pt}
\begin{tabular}{lccc}
\toprule
\textbf{Task} & \textbf{Fixed camera} & \textbf{Egocentric} & $\Delta$ \\
\midrule
Open Drawer & 46 & 88 & 42 \\
Hang Tape & 24 & 80 & 56 \\
Hang Cup & 28 & 82 & 54 \\
Pour Water & 16 & 76 & 60 \\
PnP-Box & 30 & 82 & 52 \\
Multi-PnP-Box & 12 & 74 & 62 \\
\midrule
\textbf{Avg.} & 26 & 80.3 & 54.3 \\
\bottomrule
\end{tabular}
\caption{Ablation on camera viewpoint. We train two policies on the same set of 50 demonstrations with no data augmentation, using either a fixed external RGB-D camera or an egocentric RGB-D camera mounted on the end-effector. We report per-task success rates (\%); $\Delta$ denotes the absolute improvement of the egocentric setting over the fixed-camera setting.}
\label{tab:ablation_viewpoint}
\end{table}

Table~\ref{tab:ablation_viewpoint} compares policies trained with matched demonstrations but different observation viewpoints.
The egocentric setting consistently outperforms the fixed-camera setting by a large margin across all six tasks, yielding a substantially higher average success rate.
The performance gap is most pronounced on tasks that require precise close-range interactions and involve frequent self-occlusions during grasping and placement.

\paragraph{Analysis.}
The results indicate that the advantage of the egocentric viewpoint stems from its ability to \emph{actively correct} observation distribution shift during execution. When the policy encounters OOD inputs, the motion field can move the end-effector to reframe the scene in an object-centric manner, bringing the target back to a canonical configuration in the egocentric camera view. This effectively steers the incoming observations back into the training support of the egocentric execution policy, which helps the system recover and continue the task.

By contrast, the fixed-camera policy operates on a viewpoint-absolute global observation that cannot be altered by the robot’s motion. Although the motion field can still drive the gripper toward an object-centric state that is familiar from the demonstrations, the object’s appearance in the fixed camera (e.g., its image location, scale, and visibility pattern) remains largely dictated by the static viewpoint and therefore cannot be steered toward the in-distribution region. As a consequence, the fixed-camera policy is more likely to remain exposed to OOD observations, which is consistent with its substantially lower success rates observed in Table~\ref{tab:ablation_viewpoint}.


\subsubsection{Summary}
Overall, the ablations highlight the importance of egocentric design for robust multi-stage manipulation. 
Egocentric data augmentation substantially improves performance, particularly on long-horizon tasks, by broadening the training support and increasing tolerance to motion-field pose estimation errors and observation drift. 
Moreover, egocentric sensing significantly outperforms a fixed-camera viewpoint, as the motion field can actively reframe egocentric observations back into an object-centric in-distribution regime, whereas fixed global observations cannot be similarly steered.

\subsection{Cross-Object Generalization Evaluation.}
\begin{figure}[t]
    \centering
    \includegraphics[width=0.85\linewidth]{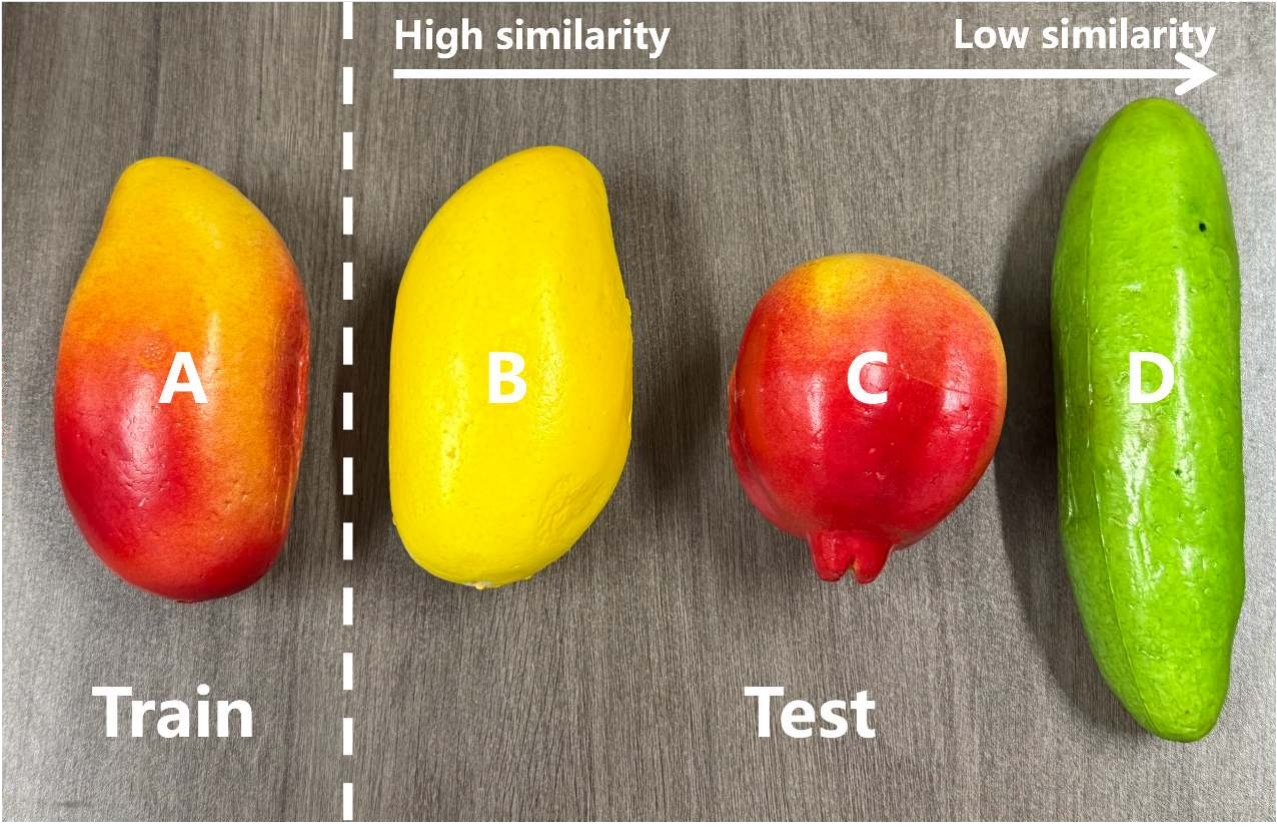}
    \caption{
    \textbf{Cross-object generalization on PnP-Box.}
    SID is trained with demonstrations using object A and evaluated on unseen test objects B--D in the PnP-Box task.
    The test objects are arranged from high to low visual similarity with respect to the training object A, covering variations in geometry and appearance.
    }
    \label{fig:cross_obj}
\end{figure}

We further evaluate SID's ability to generalize across object instances in the PnP-Box task.
As shown in Fig.~\ref{fig:cross_obj}, SID is trained using demonstrations with object A and tested on objects B--D, which are arranged from high to low visual similarity with respect to the training object.
The task structure is kept unchanged: the robot must open the box, pick the target object, and place it into the box.
Therefore, this experiment isolates the effect of object-level appearance and geometry shift while keeping the surrounding manipulation routine fixed.

As shown in Table~\ref{tab:cross_object_pnpbox}, performance decreases as the test object becomes less visually and geometrically similar to the training object.
SID retains partial transfer to visually similar unseen objects, but the success rate drops under larger changes in shape, size, and appearance.

This trend is consistent with the roles of the two SID components.
When upstream segmentation and pose estimation remain reliable, the motion field can still guide the end-effector toward a similar pre-grasp distribution across object instances, while the PointNet-style visuomotor encoder provides some robustness to appearance variations in segmented egocentric point clouds. However, large geometry or size changes can make the segmented point-cloud observations substantially different from those seen during training, reducing the reliability of the learned local actions. 

\begin{table}[t]
\centering
\small
\renewcommand{\arraystretch}{1.08}
\begin{tabular*}{0.70\linewidth}{@{\extracolsep{\fill}}lcccc@{}}
\toprule
\textbf{Method} & \textbf{A} & \textbf{B} & \textbf{C} & \textbf{D} \\
\midrule
SID-O$^{\dagger}$ & 20/25 & 14/25 & 9/25  & 4/25 \\
SID-C$^{\dagger}$ & 22/25 & 17/25 & 11/25 & 6/25 \\
\bottomrule
\end{tabular*}
\caption{
\textbf{Cross-object generalization on PnP-Box.}
SID is trained with object A and evaluated on objects A--D, where B--D are unseen test objects.
We report successful trials over 25 evaluations.
}
\label{tab:cross_object_pnpbox}
\end{table}

\subsection{Effect of the Number of Demonstrations.}
We use two demonstrations in the main experiments because a single demonstration can lead to ambiguous motion-field behavior under certain object-pose variations.
With only one demonstration, the motion field may map both the original object state and a 180$^\circ$-rotated object state to the same demonstrated manifold.
Although the resulting trajectories can still complete the task, they are often less natural and less efficient, with noticeable detours before reaching the execution policy's reliable operating region.
This explains why two demonstrations provide a more stable default setting for SID.

We further evaluate how SID-open performs as the number of raw demonstrations increases.
As shown in Table~\ref{tab:demo_num}, SID-open already achieves strong performance with two raw demonstrations after egocentric augmentation.
Increasing the number of raw demonstrations consistently improves performance across the evaluated tasks, but the improvement is modest.
In particular, increasing the number of raw demonstrations from 10 to 50 yields only small additional gains.
These results suggest that, after egocentric augmentation, SID-open can achieve strong performance from only two raw demonstrations, and the marginal benefit of adding more raw demonstrations becomes relatively small in these evaluated tasks.

\begin{table}[t]
\centering
\small
\renewcommand{\arraystretch}{1.10}
\setlength{\tabcolsep}{5pt}
\begin{tabular*}{0.76\linewidth}{@{\extracolsep{\fill}}l|ccc@{}}
\toprule
\textbf{Task} & \textbf{2 demos} & \textbf{10 demos} & \textbf{50 demos} \\
\midrule
Hang Tape & 19/25 & 20/25 & 22/25 \\
Hang Cup  & 21/25 & 20/25 & 23/25 \\
PnP-Box   & 20/25 & 22/25 & 21/25 \\
\bottomrule
\end{tabular*}
\caption{
\textbf{Effect of the number of raw demonstrations.}
We evaluate SID-open with 2, 10, and 50 raw demonstrations on three representative tasks.
All settings are augmented to 100 training trajectories using the same egocentric augmentation pipeline.
We report successful trials over 25 evaluations.
}
\label{tab:demo_num}
\end{table}

\subsection{Alternative Motion-Field Formulations.}
We study alternative formulations for the pre-policy distribution-alignment stage by comparing the proposed motion field with two explicit pose-based alignment variants.
The comparison is conducted on three representative tasks, Hang Tape, Hang Cup, and PnP-Box, using two raw demonstrations per task.
All variants use the same execution policy and differ only in the alignment module before policy execution. 

In Align-O, the robot estimates the current object pose once and moves directly to the corresponding demonstration pose.
In Align-C, the robot continuously re-estimates the object pose during alignment and repeatedly corrects its motion toward the same fixed demonstration correspondence.
Thus, Align-C can be viewed as an explicit closed-loop alignment variant when object poses and demonstration correspondences are manually available.

As shown in Table~\ref{tab:alternative_field}, Align-C improves over Align-O, indicating that online pose updates are important for reliable alignment.
Its performance is close to the learned motion field on these tasks, suggesting that explicit closed-loop alignment can be effective when reliable pose estimates and manually specified correspondences are available.
However, Align-C depends on a hand-designed pose-space construction and a fixed alignment target.
In contrast, the motion field provides a learned formulation of distribution alignment that retains online correction while avoiding manual specification of a single target correspondence at test time. This also makes the formulation less tied to this particular explicit-alignment heuristic and potentially applicable to other object-centric representations.

\begin{table}[t]
\centering
\small
\renewcommand{\arraystretch}{1.10}
\setlength{\tabcolsep}{5pt}
\begin{tabular*}{0.76\linewidth}{@{\extracolsep{\fill}}l|ccc@{}}
\toprule
\textbf{Task} & \textbf{Align-O} & \textbf{Align-C} & \textbf{Motion Field} \\
\midrule
Hang Tape & 15/25 & 19/25 & 20/25 \\
Hang Cup  & 16/25 & 22/25 & 22/25 \\
PnP-Box   & 19/25 & 22/25 & 21/25 \\
\bottomrule
\end{tabular*}
\caption{
\textbf{Comparison with explicit alignment variants.}
We evaluate SID-open and two pose-based alignment variants on three representative tasks using two raw demonstrations per task.
Align-O performs one-shot alignment, while Align-C performs closed-loop alignment with continuously updated pose estimates.
We report successful trials over 25 evaluations.
}
\label{tab:alternative_field}
\label{tab:demo_num}
\end{table}

\subsection{Extended Experiment: Make Coffee}
\begin{figure*}[t]
    \centering
    \includegraphics[width=\linewidth]{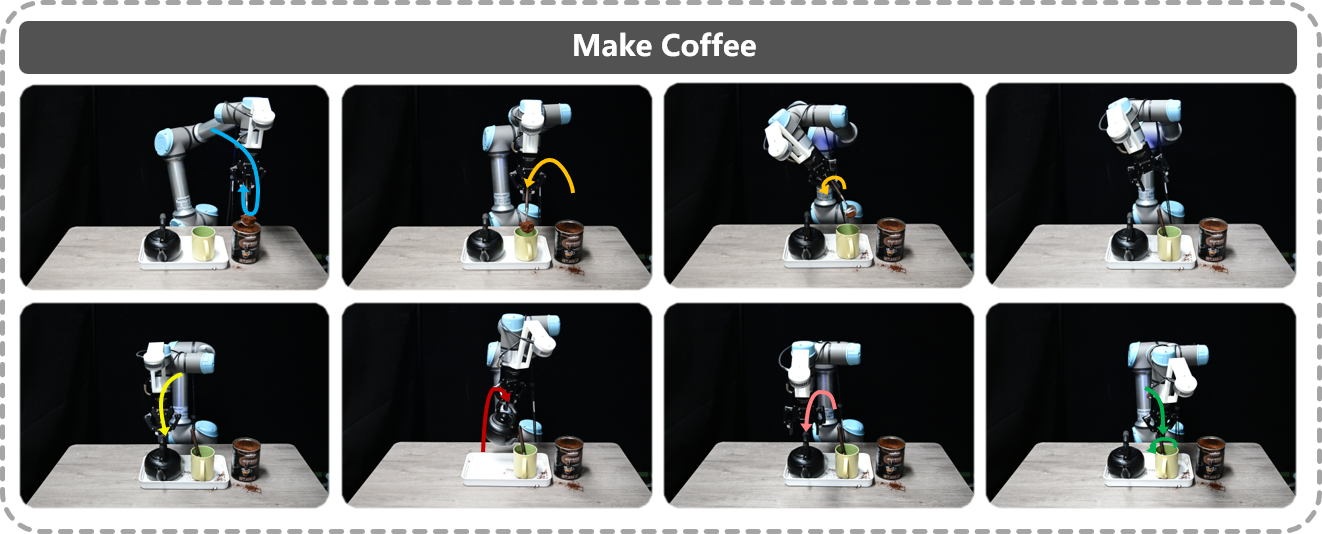}
    \caption{
    Visualization of \textsc{Make Coffee}. 
    }
    \label{fig:make_coffee}
\end{figure*}

While our previous long-horizon evaluation involves a 7-stage task, it is largely composed of repeated pick-and-place behaviors, resulting in a relatively homogeneous action pattern.
To further probe the capability of SID on more challenging long-horizon manipulation with richer skill composition and more diverse stage transitions, we conduct an extended experiment on a \textsc{Make Coffee} task.

\paragraph{Task description.}
\label{sec:make coffee discription}
The goal is to complete a full coffee-making routine from a reset tabletop configuration, requiring the system to compose heterogeneous subtasks with diverse interaction modes.
Concretely, the task is decomposed into the following stages: (i) grasp the spoon from the coffee jar, (ii) pour coffee powder into the cup and place the spoon into the cup, (iii) grasp the teapot, (iv) pour water into the cup, (v) return the teapot to the tray, and (vi) grasp the spoon and stir the coffee.

\paragraph{Evaluation protocol.}
We evaluate SID on the coffee-making task under the same hardware setup and metrics as in the main experiments.
We collect only two expert demonstrations for this task and evaluate the resulting system over 25 trials.

\paragraph{Results.}
Table~\ref{tab:coffee_stage_success} summarizes the quantitative results on the extended task. Despite the increased task complexity and the inclusion of more fine-grained interaction primitives, our method achieves a non-trivial overall success rate, demonstrating its potential for long-horizon, multi-stage manipulation.
However, the task-level success rate is noticeably lower than that on \textsc{Multi-PnP-Box}, which also consists of seven stages.
This gap suggests that, while the proposed pipeline scales to long horizons, further improvements are needed to better handle more precise and contact-rich manipulation subtasks.

\begin{table}[t]
\centering
\scriptsize
\setlength{\tabcolsep}{4.0pt} 
\renewcommand{\arraystretch}{1} 
\begin{tabular}{@{}lccccccc@{}}
\toprule
 & \textbf{S1} & \textbf{S2} & \textbf{S3} & \textbf{S4} & \textbf{S5} & \textbf{S6} & \textbf{Overall} \\
\midrule
\textbf{Success / 25} & 24/25 & 20/25 & 20/25 & 16/25 & 14/25 & 14/25 & 14/25 \\
\bottomrule
\end{tabular}
\caption{Per-stage success counts on the extended \textsc{Make Coffee} task over 25 trials. S1--S6 correspond to the six stages described before.}
\label{tab:coffee_stage_success}
\end{table}

\subsection{Failure Case Analysis}
\label{sec:failure_analysis}

We summarize typical failure cases observed in our real-world evaluation. In most instances, failures stem from imperfect perception and limited scene context, rather than systematic breakdowns of the control components.

\textbf{Limited wrist-view context.}
Because perception is driven by a wrist-mounted camera, the available visual context can be constrained by field of view, self-occlusion, and viewpoint geometry. A simple mitigation is to move the end-effector to an elevated ``look'' pose before execution, which often improves visibility and stabilizes downstream perception. Nonetheless, some tabletop scenes remain visually ambiguous even from a higher viewpoint (e.g., small items, low-texture objects, or visually similar instances), which can occasionally confuse target identity. When this occurs, target isolation becomes less reliable and can affect both object-centric state construction and policy inputs.

\textbf{Segmentation imperfections.}
Our pipeline relies on target isolation to construct both the object-centric state for the motion field and the egocentric observation for the policy. In cluttered scenes or under partial occlusions, segmentation may return incomplete masks, include small background fragments, or miss thin structures (e.g., handles or tape edges). While the system is generally tolerant to mild noise, more pronounced mask errors can degrade the segmented point cloud used by the field and the policy.

\textbf{Pose estimation instability.}
The motion-field module uses an auxiliary object pose estimate to canonicalize pre-contact states. Under challenging viewpoints, reflective surfaces, texture-poor objects, or heavy occlusion, pose estimation can occasionally become less stable (e.g., jitter or intermittent hypothesis switches). This issue can be further exacerbated in more dynamic situations, such as rapid wrist motion (causing blur or large viewpoint change) or noticeable object motion during execution, where the pose tracker may momentarily lag or produce inconsistent updates. Reduced pose stability weakens object-centric alignment, leading to less accurate field updates and slower or imperfect convergence toward the demonstrated pre-contact manifold.

\textbf{Target-centric perception with limited environment awareness.}
Both the field and the policy primarily operate on target-centric observations. As a result, nearby obstacles and supporting surfaces (e.g., container rims, shelf walls, table edges) may be under-represented when they fall outside the wrist view or are not included by target isolation. In constrained workspaces or cluttered setups, this can occasionally cause unintended contacts or collisions with the environment during approach or while executing corrective motions, even when the target object itself is correctly perceived.

\begin{figure*}[t]
    \centering
    \includegraphics[width=\linewidth]{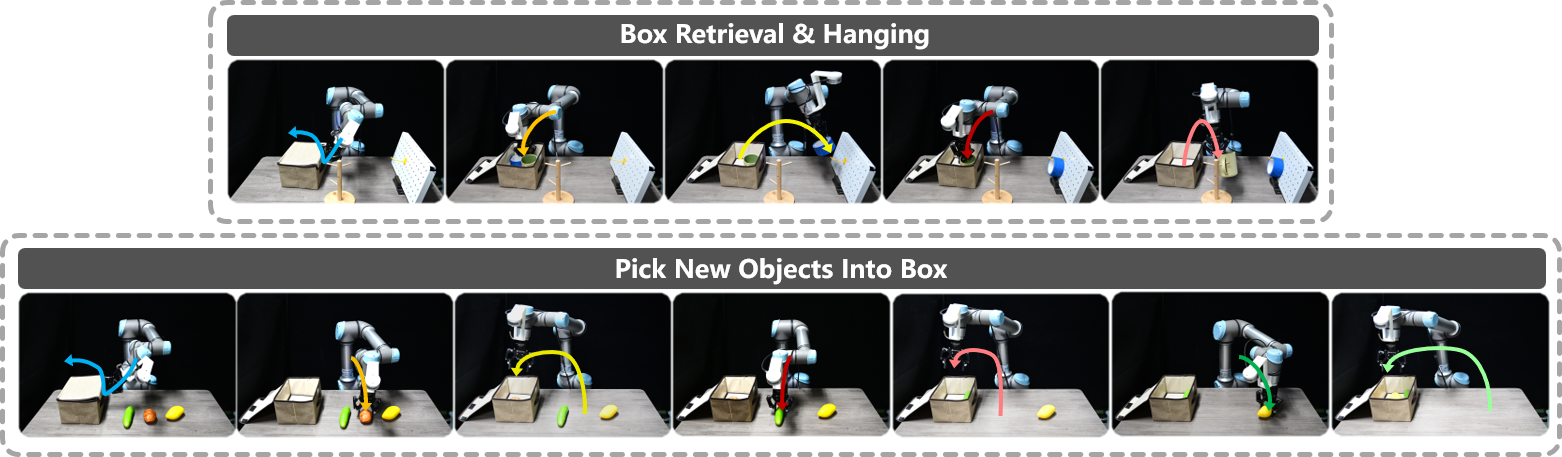}
    \caption{
    Visualization of the recomposed tasks. 
    }
    \label{fig:recompoes_exp}
\end{figure*}

\subsection{Additional Figures}
\label{sec:additional_figures}

Fig.~\ref{fig:recompoes_exp} illustrates the recomposed task variants used to evaluate skill reuse and composition.

\section*{Appendix B}

\subsection{Mutil-stage Task Processing}
For multi-stage tasks involving multiple objects, SID decomposes each task into a sequence of stages, where a single object is selected as the target for processing at each stage. Fig~\ref{fig:task_decomposition} presents the decomposition of each mentioned task into its corresponding stages and associated objects.

\textbf{Stage-aware Augmentation for Multi-stage Tasks.}
For multi-stage demonstrations, we first run a heuristic keyframe detection procedure over the full trajectory.
Specifically, we identify keyframes by jointly thresholding (i) the magnitude of pose change along the trajectory and (ii) the gripper aperture variation.
This produces a sparse set of keyframes that delineates both stage transition boundaries and the onset/termination of reaching within each stage.
We then select the approach segments defined by these keyframes and apply augmentation only within these segments. Each augmented segment is attached with a 

\textbf{Stage Transition in Inference Time.}
During inference, SID performs stage transitions based on a termination signal produced by the egocentric execution policy. 
Specifically, at time step $t$, the policy outputs a termination score $\textsc{TermDim}(A_t)$, which is used to decide whether the current stage should end.

We denote an ordered sequence of $N$ stages (subtasks) as
\begin{equation}
\mathbf{S} = \{ \mathrm{subtask}_i \}_{i=1}^{N}.
\end{equation}
Correspondingly, we maintain a key sequence
\begin{equation}
\mathbf{K} = \{ (k^{i}_{\mathrm{obj}},\, k^{i}_{\mathrm{task}}) \}_{i=1}^{N},
\end{equation}
where $(k^{i}_{\mathrm{obj}}, k^{i}_{\mathrm{task}})$ specifies the object key and task key associated with $\mathrm{subtask}_i$.

While executing $\mathrm{subtask}_i$, the system monitors $\textsc{TermDim}(A_t)$. 
Once $\textsc{TermDim}(A_t) > \epsilon_{\mathrm{term}}$, we declare $\mathrm{subtask}_i$ as terminated. 
The robot then returns to an initial observation pose $T_{\mathrm{init}}$, and the active keys are advanced to the next stage,
\begin{equation}
(k_{\mathrm{obj}}, k_{\mathrm{task}}) \leftarrow (k^{i+1}_{\mathrm{obj}},\, k^{i+1}_{\mathrm{task}}).
\end{equation}

\subsection{Convergence Intuition and Capture Region.}
The motion field is best viewed as an approximate descent system over the bounded object-centric pose domain $\Omega$ used for field training.
Using the potential $E(x)=\frac{1}{2}d(x,\mathcal X_{\mathrm{demo}})^2$, the ideal target field $v^*(x)$ points toward the nearest demonstrated approach state with a distance-scheduled magnitude that vanishes near the demonstrated manifold.
Therefore, with sufficiently small execution steps, the ideal sliding dynamics monotonically reduce the distance to $\mathcal X_{\mathrm{demo}}$ until reaching its neighborhood.
In practice, the learned field only approximates this descent direction, so convergence is empirical and local: SID can recover from perturbations that remain inside the sampled pose domain $\Omega$, with reliable pose estimation and reachable corrective motions.
Once perturbations move the object-centric state outside this domain, or make the approach manifold unobservable or unreachable, the state falls outside the capture region and the field no longer provides a convergence guarantee.

\subsection{Generalization Analysis}
\label{sec:gen_analysis}

We analyze SID along one practical axis: workspace-wide spatial generalization.
We focus on concrete design choices and the contribution of each module. 

\textbf{Workspace-wide spatial generalization.}
\emph{Base-free, egocentric policy interface.}
The execution policy conditions only on wrist-centric observations and predicts relative end-effector motions, without using any base-frame states or global coordinates.
This makes the learned behavior invariant to absolute placement in the workspace and avoids coupling to a particular robot base calibration.

\begin{figure}[t]
    \centering
    \includegraphics[width=\linewidth]{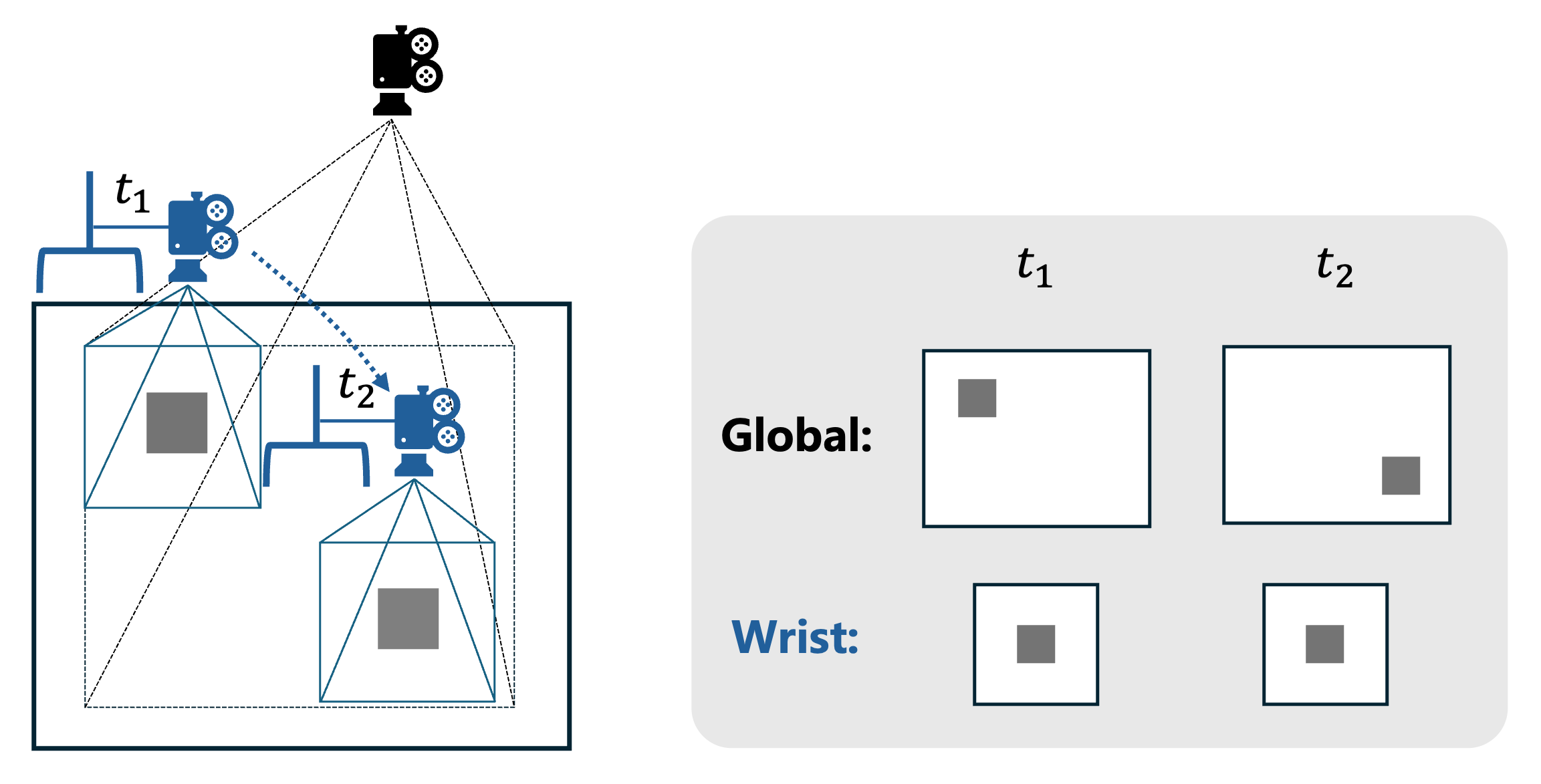}
    \caption{
    Illustration of workspace-wide generalization with an egocentric (wrist-centric) interface.
    When the object is placed at different workspace locations ($t_1$ vs.\ $t_2$), the \textbf{global} camera view changes substantially (e.g., object position in the image shifts), whereas the \textbf{wrist} view remains comparatively consistent around the interaction geometry.
    This motivates conditioning the execution policy on wrist-centric observations and predicting relative end-effector motions, enabling a shared in-distribution operating region across the workspace.
    }
    \label{fig:wrist_global_generalization}
\end{figure}

\emph{Why wrist-only is essential for the field--policy pipeline.}
SID relies on using the motion field to bring the current state into a region where the execution policy is reliable.
As illustrated in Fig.~\ref{fig:wrist_global_generalization}, the global view can vary substantially across workspace locations, while the wrist-centric view remains comparable across placements, making ``in-distribution'' checks and the field-to-policy handoff well-defined.
This requires that ``being in-distribution'' is determined by quantities that remain comparable across workspace locations.
If the policy (or its training distribution) depended on base-frame states, then the same local interaction geometry could correspond to very different global coordinates across the workspace, breaking the notion of a shared operating region and making the field-to-policy handoff ill-posed.
By defining both the policy input and actions in the wrist frame, the field correction and policy execution are aligned through a consistent egocentric interface, so a single demonstration-derived operating region can be reused across workspace-wide object placements.

\emph{Egocentric reprojection augmentation for local robustness.}
Point-cloud reprojection produces kinematically consistent viewpoint perturbations in the wrist frame.
This expands the policy's tolerance to moderate viewpoint and pose changes around the in-distribution region, improving success after the field-guided entry.

\subsection{Robustness Analysis}
\label{sec:robust_analysis}

We analyze SID's robustness along two practical axes: (i) robustness to disturbances and (ii) robustness to distractors/clutter.
We focus on concrete design choices and the contribution of each module.

\textbf{Robustness to disturbances.}
\emph{Field feedback yields disturbance tolerance even in open-loop.}
Although the open-loop pipeline does not alternate between modules, the motion field itself is a feedback mechanism: it continuously evaluates the current pre-contact state and produces a corrective update whose magnitude reflects deviation from the demonstrated manifold (e.g., via field norm/energy).
As a result, if a disturbance displaces the object or shifts the wrist viewpoint during pre-contact, the field naturally increases its corrective action and keeps sliding the system back toward the manifold until a convergence criterion is met.

\emph{Closed-loop recovery with policy-side in-distribution monitoring.}
In the closed-loop pipeline, once the system enters interaction and the execution policy takes over, we additionally equip the policy with an in-distribution confidence $p_{\text{ID}}$.
When $p_{\text{ID}}$ indicates that the current condition has drifted outside the policy's reliable region, control returns to the pre-contact correction routine and the field is re-invoked to re-enter the manifold before policy execution resumes.
This provides a second, policy-triggered recovery path beyond the field's own convergence feedback.

\textbf{Robustness to distractors and clutter.}
\emph{Target isolation for the field.}
The motion field operates on an object-centric state constructed from target-object perception (segmentation and, optionally, pose estimation), which suppresses irrelevant geometry from distractors.

\emph{Target-centric observation for the policy.}
The policy conditions on segmented point clouds of the target object together with the gripper geometry (and gripper width) in the wrist frame.
This reduces dependence on background layout and lowers sensitivity to arbitrary distractor placement, provided the target object remains at least partially observable.

\subsection{Details of Upstream Visual Modules.}
We construct the visual perception module using off-the-shelf components to obtain
temporally consistent object masks and object poses.

In the first frame of each episode, we run SAM3 with text prompts to generate
instance masks, and select the target instance according to the task specification.
The selected mask serves as the initialization for the mask tracker. For subsequent
frames, we use XMem to propagate the target mask over time. By maintaining an
explicit memory of past appearances, XMem provides a temporally consistent mask
sequence under moderate viewpoint changes and partial occlusions.

To reduce tracking drift in challenging sequences, we optionally re-run SAM3 to
re-initialize the tracker when the propagated mask shows low confidence or abrupt
changes. This simple re-initialization strategy improves long-horizon robustness
without introducing task-specific training.

The propagated mask is further used to interface the segmentation module with
FoundationPose. Given the RGB-D observation, the target mask, and the object
geometry mesh, FoundationPose estimates the object pose in the camera frame. The
mask suppresses background clutter and defines the object support region, while the
estimated pose provides a metric object-centric representation for downstream
manipulation.

\begin{figure}[t]
    \centering
\includegraphics[width=1\linewidth]{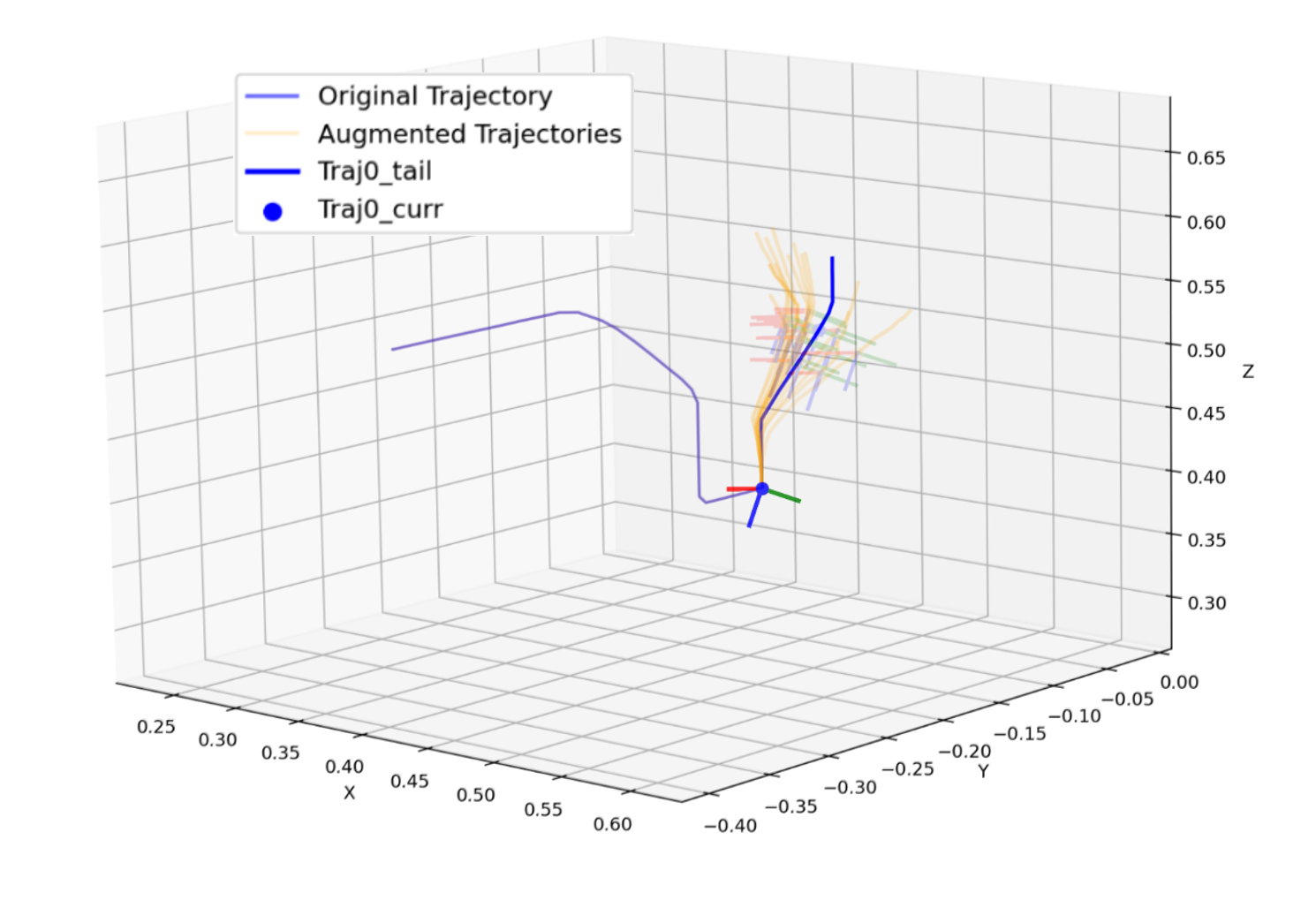}
    \caption{Illustration of augmented trajectories for \textsc{Open Drawer}, visualized in the robot base frame. As shown, we apply a decaying augmentation scheme only to certain segment of the trajectory, corresponding to the approach-to-drawer phase.}

    \label{fig:augmentation}
\end{figure}

\subsection{Details of Ego-centric Augmentation}
Specifically, for in-distribution data, we design an augmentation distribution that gradually contracts toward the demonstration manifold. Concretely, we apply a time-decaying perturbation along each demonstration trajectory: we inject larger perturbations at the beginning of the trajectory and progressively reduce the perturbation magnitude over time, eventually recovering the original trajectory. Figure~\ref{fig:augmentation} illustrates the resulting schedule and examples of the augmented trajectories.

For out-of-distribution data, we apply a bounded random perturbation to each point along the original trajectory, with lower and upper magnitude limits chosen to ensure that the perturbed states lie outside the augmented in-distribution data support.


\end{document}